\documentclass[10pt,journal,compsoc]{IEEEtran}

\usepackage{graphicx}
\usepackage{amsfonts}
\usepackage{amsmath}
\usepackage{amssymb}
\usepackage{diagbox}
\usepackage{multirow}
\usepackage{epstopdf}
\usepackage{tabularx}
\usepackage{epsfig}
\usepackage{color,subfigure}
\usepackage{times}
\usepackage{mathrsfs}
\usepackage{bm}
\usepackage{algorithm, algorithmic}
\usepackage{mathdots}
\usepackage[american]{babel}
\usepackage{microtype} 
\usepackage{array}
\usepackage{setspace}
\usepackage{threeparttable}
\usepackage{url}
\usepackage[colorlinks,linkcolor=red]{hyperref}
\usepackage{booktabs}
\usepackage{diagbox}
\usepackage{wrapfig}
\usepackage{float}
\usepackage{amsthm}
\usepackage{mathrsfs}

\ifCLASSOPTIONcompsoc

  \usepackage[nocompress]{cite}
\else
  \usepackage{cite}
\fi

\ifCLASSINFOpdf
\else
\fi

\begin{document}

\title{Simultaneously Optimizing Perturbations and Positions for Black-box Adversarial Patch Attacks}

\author{Xingxing Wei, \emph{Member, IEEE},
        Ying Guo,
        Jie Yu,
        and~Bo Zhang
\IEEEcompsocitemizethanks{
\IEEEcompsocthanksitem Xingxing Wei, Ying Guo and Jie Yu are with the Institute of Artificial Intelligence, Beihang University, No.37, Xueyuan Road, Haidian District, Beijing, 100191, P.R. China. Bo Zhang is with Tencent Corporation.
(E-mail: \{xxwei, yingguo, jieyu\}@buaa.edu.cn, cradminzhang@tencent.com)
\IEEEcompsocthanksitem Xingxing Wei is the corresponding author}
}


\IEEEtitleabstractindextext{%
\begin{abstract}
   Adversarial patch is an important form of real-world adversarial attack that brings serious risks to the robustness of deep neural networks. Previous methods generate adversarial patches by either optimizing their perturbation values while fixing the pasting position or manipulating the position while fixing the patch's content. This  {reveals} that the positions and perturbations are both important to the adversarial attack. For that, in this paper, we propose a novel method to \textbf{simultaneously} optimize the position and perturbation for an adversarial patch, and thus obtain a high attack success rate in the black-box setting. Technically,  {we regard the patch's position, the pre-designed hyper-parameters to determine the patch's perturbations as the variables}, and utilize the reinforcement learning framework to \textbf{simultaneously} solve  for the optimal solution based on the rewards obtained from the target model with a small number of queries. Extensive experiments are conducted on the Face Recognition (FR) task, and results on four representative FR models show that our method can significantly improve the attack success rate and  query efficiency. Besides, experiments on the commercial FR service and physical environments confirm its practical application value.   {We also extend our method to the traffic sign recognition task to verify its generalization ability.}
\end{abstract}

}

\maketitle

\IEEEdisplaynontitleabstractindextext

\IEEEpeerreviewmaketitle

\IEEEraisesectionheading{\section{Introduction}\label{sec:introduction}}

\IEEEPARstart{D}{eep} neural networks (DNNs) have shown excellent performance in many tasks \cite{guo2020deep}, \cite{minaee2021image}, \cite{baltruvsaitis2018multimodal}, but they  {are vulnerable} to adversarial examples \cite{szegedy2013intriguing}, where adding small imperceptible perturbations to the image  {could confuse the network}. However,  {this form of attack} is not suitable for real applications  {because images are captured through the camera and the perturbations also need to be captured}. An available way is to use a local patch-like perturbation, where the  {perturbation's magnitudes are} not restricted. By printing out the adversarial patch and pasting it on the object, the attack in the real scene can be realized.  {Adversarial patch \cite{brown2017adversarial}} has brought security threats to many tasks like  traffic sign recognition \cite{rp2,ganpatch}, image classification \cite{karmon2018lavan, wang2020hamiltonian}, as well as person detection and re-identification \cite{tshirt, bai2020adversarial}.
 
Face Recognition (FR) is  {a relatively  safety-critical task}, and the adversarial patch has also been successfully applied in this area \cite{xiao,advhat,glass2019,advsticker}. For example, adv-hat \cite{advhat} and adv-patch \cite{patch,xiao,xiao2021improving} put the patch generated based on the gradients on the forehead, nose, or eye area to achieve attacks. Adv-glasses \cite{glass2016,glass2019} confuse the FR system by placing a printed perturbed eyeglass frame at the eye. The above methods mainly focus on optimizing the patch's perturbations, and the patch's pasting position is fixed on a  {location} selected based on  the prior knowledge. On the other hand, adv-sticker \cite{advsticker} adopts a predefined meaningful adversarial patch,  {and} uses an evolutionary algorithm to search for a good patch's position to perform the attack, which shows  {that} the patch's position is one of the major parameters on the patch attacks. The above methods  {motivate} us that if the position and perturbation are optimized at the same time,  {the high} attack performance could be achieved. 

However, simultaneous optimization cannot be  viewed as a  {simple} combination of these two independent factors.  There is a strong  {coupling relationship}  between the position and perturbation. Specifically, experiments show that the perturbations generated at different positions on the face tend to resemble the facial features of the current facial region (detailed in Section \ref{sec:3.1}). This verifies that the two factors are interrelated and  {influenced with} each other.  {Therefore, they}  should be optimized simultaneously, which means a simple two-phase method or an alternate iterative optimization is not optimal.  Additionally, in practical applications, the detailed information about the target FR model usually cannot be accessed.  {Instead, some commercial online vision APIs (e.g. Face++ and Microsoft cloud services) usually return the predicted identities and scores for the uploaded face images. By utilizing this limited information, exploring the query-based black-box attack to construct  adversarial patches is a reasonable solution. In such a case, the simultaneous optimization will lead to a large searching space, and further bring a lot of queries  to the FR system. Thus, \emph{how to design an efficient simultaneous optimization mechanism to solve for these two factors in the black-box setting becomes a challenging problem for further study.}}

\begin{figure*}[t]
\centering
\includegraphics[width=0.95\textwidth]{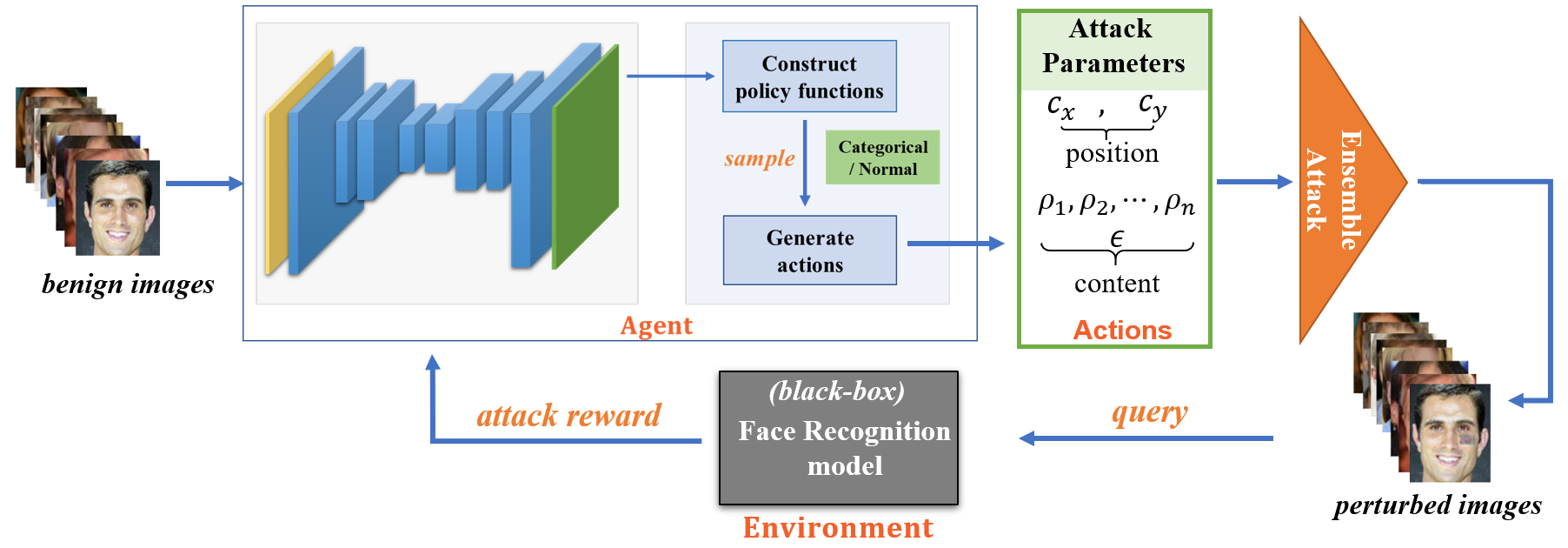}
\caption{ An overview of simultaneously optimizing positions and perturbations based on the reinforcement learning (RL) framework.
 {The benign image is fed into the agent, and the policies in RL are constructed by the agent. After sampling according to the policies, the specific solutions for the attack variables (actions) are determined. By the ensemble attack and querying the target model, rewards can be obtained, and the parameters of the agent are updated in the iteration.}}
\label{fig:frm3}
\end{figure*}

 {Currently, some works \cite{rao2020adversarial, yang2020patchattack, xiang2021gdpa} have studied the optimization of both positions and perturbations. Location-optimization \cite{rao2020adversarial} uses an alternate iterative strategy to optimize one while fixing the other under the white-box setting.
TPA \cite{yang2020patchattack} uses reinforcement learning to search for suitable patch textures and positions in the black-box setting. It belongs to the two-phase method, and because the patch's pattern comes from a  set of predefined texture images, it requires thousands of queries to the target model. 
 GDPA \cite{xiang2021gdpa} considers the intrinsic coupling relationship, generating both of them through a generator. However, it is a white-box attack and also has other shortcomings, such as the need for the time-consuming offline training and the restrictions to only deal with continuous parameter spaces (detailed in Section \ref{sec:2.1}). Therefore, these methods cannot meet the challenge.}

Based on above considerations, in this paper, we propose an  {efficient} method to \emph{simultaneously} optimize the position and perturbation of the adversarial patch to improve the black-box attack performance with limited information. 
 {
For the perturbation, pixel values of each channel in the patch range from 0 to 255, which results in a huge searching space that is time-consuming to optimize.
To tackle this issue, we first reduce the searching dimension.
Specifically, based on the transferability of attacks \cite{demontis2019adversarial}, we utilize the modified I-FGSM \cite{dong2018boosting,ti,wang2021improving} conducted  on the ensemble surrogate models \cite{esm}, and adjust its hyper-parameters (i.e., the attack step size and the weight for each surrogate model) to generate the patch's perturbations. In this process, the black-box target model will be well-fitted by the ensemble surrogate models with a proper set of weights, and thus the computed transferable perturbations can perfectly attack the target model.} Compared with directly optimizing the pixel-wise perturbation values, changing the attack step size and models' weights can greatly reduce the parameter space and improve the solving efficiency.  {Based on this setting, the patch's position, the hyper-parameters for the patch’s perturbations (i.e., the surrogate models' weights and the attack step size) become the key variables that ultimately need to be simultaneously learned.}

To search for the optimal solution, we use a small number of queries to the target model to dynamically adjust the above learnable variables. This process can be formulated into the Reinforcement Learning (RL) framework.  {Technically, we first carefully design the actions for each variable, and then 
utilize an Unet-based \cite{unet} agent to guide this simultaneous  {optimization} by designing the policy function between the input image and the output action for each variable. The actions are finally determined according to the policy functions.}
The environment in the RL framework is set as the target FR model.
By interacting with the environment, the agent can obtain a reward signal feedback to guide its learning by maximizing this reward \cite{li2017deep}.  {Note that our method is an online learning process, we don't need to train the agent in advance. Given a benign image, the agent begins from the random parameters. The  reward in each iteration will guide the agent to become better and better  until the attack is achieved.} The whole scheme is illustrated in Figure \ref{fig:frm3}. 
The code can be found in \url{https://github.com/shighghyujie/newpatch-rl}.

The generated adversarial patch can also combine with the existing gradient estimation attack methods \cite{chen2017zoo,nes, random} in a carefully designed manner, and  provide a good initialization value for them. Thus, the high query cost of the gradient estimation process can be reduced and the attack success rate is further improved (see Section \ref{sec:extend1}). 
In addition to the face recognition, our method is also applicable to other scenarios, such as the traffic sign recognition task. The experiments are given in Section \ref{sec:traffic}.

In summary, this paper has the following contributions:
\begin{itemize}
\setlength{\itemsep}{0pt}
\item We  {empirically illustrate} that the position and perturbation of the adversarial patch are equally important and interact with each other closely. Therefore, taking advantage of the mutual correlation, an  {efficient} method is proposed to \emph{simultaneously} optimize them to generate an adversarial patch  {in the black-box setting.}
\item We formulate the optimization process into a RL framework, and carefully design the agent structure and the corresponding policy functions, so as to guide the agent to  {efficiently} learn the optimal parameters.
\item Extensive experiments in face dodging and impersonation tasks confirm that our method can realize the state-of-the-art attack performance and  {high} query efficiency (maximum success rate of 96.65\% with only 11 queries on average), and experiments in the commercial API and physical environment prove the good application value.
\item To show the flexibility of the proposed method,  we combine it with existing gradient estimation  {attack} methods to reduce their high query costs and improve the attack performance. Besides, we extend the proposed method to the traffic sign recognition task to verify its generalization ability. 
\end{itemize}

 \par The remainder of this paper is organized as follows. Section \ref{sec:2} briefly reviews the related work. We introduce the details of the proposed  adversarial patch method against face recognition task in Section \ref{sec:method}.
 Section \ref{sec:extend} gives the method about how to combine with the gradient estimation  {attacks}. Section \ref{sec:exp}  shows a series of experiments, and Section \ref{sec:traffic} gives the extensions to traffic sign recognition task.  Finally, we conclude the paper in Section \ref{sec:conclusion}.

\section{Related Work}\label{sec:2}


\subsection{Adversarial patch} \label{sec:2.1}

Compared with $L_p$-norm based adversarial perturbations, the adversarial patch \cite{brown2017adversarial} is a more suitable attack form for real-world applications where objects need to be captured by a camera. Different from pixel-wise imperceptible  {perturbations}, the adversarial patch does not restrict the perturbations' magnitude.  
Up to now, the adversarial patch has been applied to image classifiers \cite{watermark,karmon2018lavan}, person detectors \cite{tshirt}, traffic sign detectors \cite{rp2,ganpatch}, and many other security-critical systems. For example,  the adversarial T-shirt \cite{tshirt} evades the person detector by printing the patch generated by the gradients in the optimization framework on the center of the T-shirt. The work in \cite{ganpatch} confuses the traffic sign detection system by pasting the patch generated by the generative adversarial network on a prior fixed position of the traffic sign. The work in \cite{rp2} attacks the image classifier and the traffic sign detector by pasting stickers generated by the robust physical perturbation algorithm on the daily necessities and the traffic sign. Therefore, the adversarial patch has become an important method to help evaluate the robustness of DNN models deployed in the real life.

Recently, Location-optimization \cite{rao2020adversarial}  proposes to jointly optimize  {the} location and content of an adversarial patch, but there are three limitations: (1) They belong to the white-box attack for image classification, which needs to know detailed information of target classifiers. (2) They optimize  {the two factors} via alternate iterations, where one factor is fixed when the other factor is solved, so it is not the simultaneous optimization. (3) The pattern generated at a position is often more applicable to the area near this position. So after the pattern is optimized, the range of position change is limited, and the optimal position  within the entire image cannot be achieved.   {As a comparison}, our method can better meet the challenge. 

Another related paper is presented in \cite{yang2020patchattack}, where the authors parameterize the appearance of adversarial patches by a dictionary of class-specific textures, and then optimize the position and texture parameters of each patch using reinforcement learning like ours.  However, we are different from \cite{yang2020patchattack} in the following aspects:  (1) the position and texture in \cite{yang2020patchattack} are not solved simultaneously. They explore a two-step  {mechanism, which first learns} a texture dictionary and then uses RL to search  {for} the patch's texture on the dictionary as well as the position in the input image. This separate operation limits the performance. (2) Because of the defective formulation, the method in \cite{yang2020patchattack} shows the poor query efficiency, while our method only needs very few queries to finish the black-box attacks.  (3) \cite{yang2020patchattack} aims at attacking the DNN classification models, while our method can perform attacks on face recognition and traffic sign recognition. Furthermore, we show the effectiveness in the physical world. So, our method is more practical than \cite{yang2020patchattack}. 

 {Additionally, GDPA \cite{xiang2021gdpa} considers the simultaneous optimization  as we do by training a universal generator, but there are still some differences: (1) It is a white-box attack, where the training of the generator needs to obtain the gradient information of the model. While our method is a black-box attack, which is more applicable in the real life.
(2) The generator that outputs the two parameters is trained offline and can only be used for the target model and the unique target identity specified at training. When attacking other models or other identities, the generator needs to be retrained.
In contrast, our method is learned online and is not limited to  the specified model and identity in advance. 
(3) It is solved in a continuous parameter space, but in some scenarios, the valid position values may be discontinuous (e.g., in the FR system, the patch position is often required not to cover the facial features), and our method can deal with such discontinuous cases (Section \ref{sec:opt}).
(4) In this method, the motivation of simultaneous optimization for the two key factors is intuitive without a careful analysis, but we provide a more detailed analysis and some visualization results (Section \ref{sec:3.1}) to better explore this coupling relationship.}

\subsection{Adversarial patch in the face recognition}
Adversarial patches also bring risks to face recognition and detection tasks, and their attack forms can be roughly divided into two categories. On the one hand, some methods fix the patch on a specific position of the face selected based on the experience or prior knowledge, and then generate the perturbations of the patch. For example, adversarial hat \cite{advhat}, adv-patch \cite{patch}, and adversarial glasses \cite{glass2016,glass2019} are classical methods against face recognition models which are realized by placing perturbation stickers on the forehead or nose, or putting the perturbation eyeglasses on the eyes. GenAP \cite{xiao2021improving} optimizes the adversarial patch on a low dimensional manifold and pastes them on the area of eyes and eyebrows. The main concern of these methods is to mainly focus on generating available adversarial perturbation patterns but without considering the impact of patch's position versus the attack performance. 

On the other hand, some methods fix the content of the adversarial patch and search for the optimal pasting position within the valid pasting area of the face. RHDE \cite{advsticker} uses a pattern-fixed sticker existing in the real life, and changes its position through RHDE algorithm based on the idea of differential evolution to attack FR systems. We believe that the position and perturbation of the adversarial patch are equally important to attack the face recognition system, and if the two are optimized simultaneously, the attack performance can be further improved.

\subsection{Deep Reinforcement Learning}
Deep reinforcement learning (DRL) combines the perception ability of deep learning with the decision-making ability of reinforcement learning, so that the agent can make appropriate behaviors through the interaction with the environment \cite{li2017deep,dong2019attention}. It receives the reward signal to evaluate the performance of an action taken through the agent without any  {other} supervisory information, and can be used to solve multiple tasks such as parameter optimization and computer vision \cite{li2017deep}. In this paper, we apply the RL framework to solve for the attack variables, which can be formalized as the process of using reward signals to guide the agent's learning. Therefore, a carefully-designed agent is proposed to learn the parameter selection policies, and generate better attack parameters under the reward signals obtained by querying the target model.

\section{Methodology}\label{sec:method}

\subsection{The interaction of positions and perturbations}\label{sec:3.1}
To  understand the intrinsic coupling  {relationship} between positions and perturbations, we conduct experiments on 1000 face images. 

\begin{figure*}[h]
\centering
\includegraphics[width=0.8\textwidth]{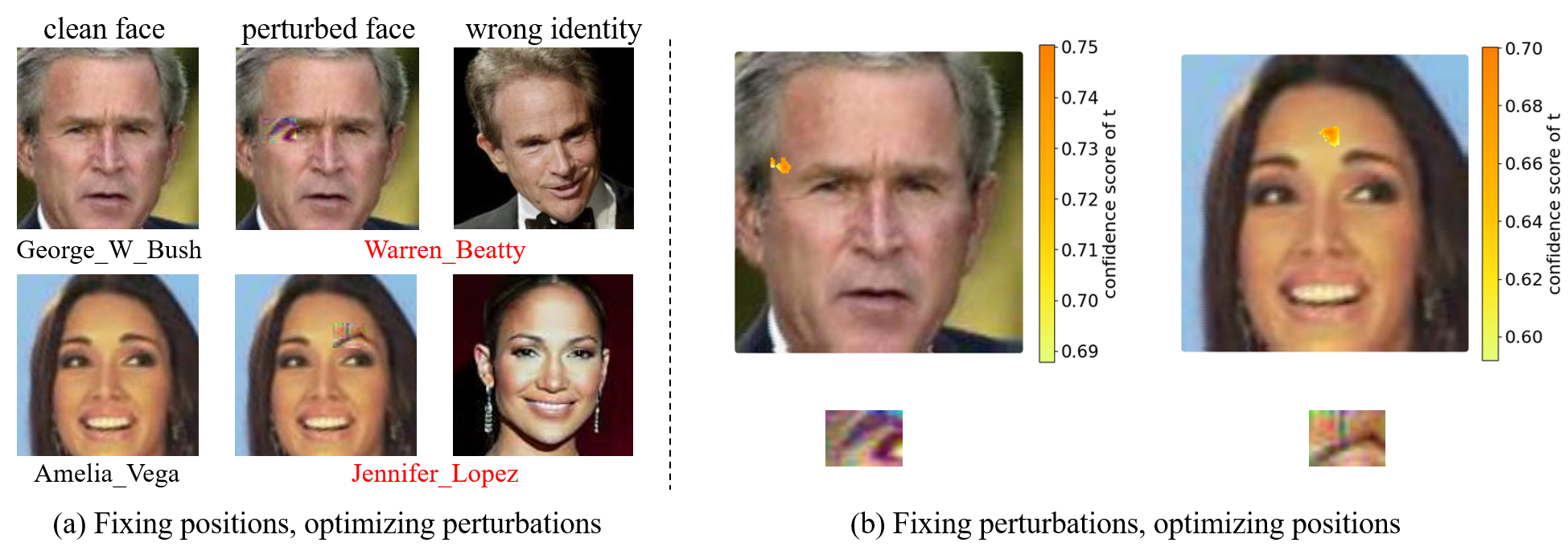}
\caption{Examples showing the interaction of positions and perturbations. In (a), we give two examples generated by fixing patch's positions while optimizing its perturbations. The black text below the images denotes the ground-truth identity, and the red text is the wrong identity after attacks.  In (b), we give two examples generated by optimizing patch's positions for a given fixed perturbation.  We show all the available positions that can lead to a successful attack in the face images (the orange region denotes all the available patch's center points), the $t$ in the y-axis is the wrong identity. }
\label{fig:visb}
\end{figure*}

Firstly, we paste patches on fixed positions in different facial regions to see the differences between their  {perturbations}. To make the effect more obvious, we choose the positions close to the key facial features (i.e. eyes, eyebrows, nose, mouth, etc.), and allow a small range of coverage (no more than 30 pixels) of the facial features. Two examples  {generated by MI-FGSM} \cite{dong2018boosting} are shown in Figure \ref{fig:visb} (a). Interestingly, we find that the perturbations generated at different positions on the face tend to resemble the facial features of the current facial region, but their shapes have been changed. For example, in the top row  of Figure \ref{fig:visb} (a), the patch is pasted to the area near the left eye, and the generated pattern is also like the shape of the eye, but it is different from the real identity's eye; in the bottom row of Figure \ref{fig:visb} (a), the adversarial patch is pasted on the left side eyebrow, and the generated pattern is similar to the shape of the eyebrow, which changes the eyebrow's shape, and may mislead the extraction of eyebrow features by the face recognition network. These phenomena indicate that the pattern of the patch is strongly correlated with the pasting position. Patterns  tend to be the features of the face region in which they are located, and the pattern of the patch varies greatly from region to region.

Then, we  study the patch's pasting position with a given pattern to the adversarial attacks. For the patch patterns shown in Figure \ref{fig:visb} (a), we exhaustively search for their positions (i.e., the patch's center points) that can realize successful attacks to the target wrong identity within the range of valid face pasting regions (see Figure \ref{fig:visb} (b)), and find that for these given patterns, the available positions (the orange region in the faces) are still concentrated near their original pasting positions. In other words, patterns generated at one position tend to have good attack effects only in a small area near this position. Besides,  we find that the confidence score for the target wrong identity will decrease when the patch's position is gradually moving away from its original position (the color denotes the score). This shows the patch's optimal position is closely relevant to its pattern, and changing the patch's position will lead to different adversarial attack effects.

In conclusion, a simultaneous optimization for these two factors is needed to optimize the matching of the position and perturbation that can realize the attack in the \emph{entire} parameter space.  {However, as discussed above, the simultaneous optimization will lead to a large searching space, and further bring a lot of queries to the FR system in the black-box attacks.}  So an efficient simultaneous optimization is a challenging problem.

\subsection{Problem formulation}

\begin{figure*}[t]
\centering
\includegraphics[width=0.95\textwidth]{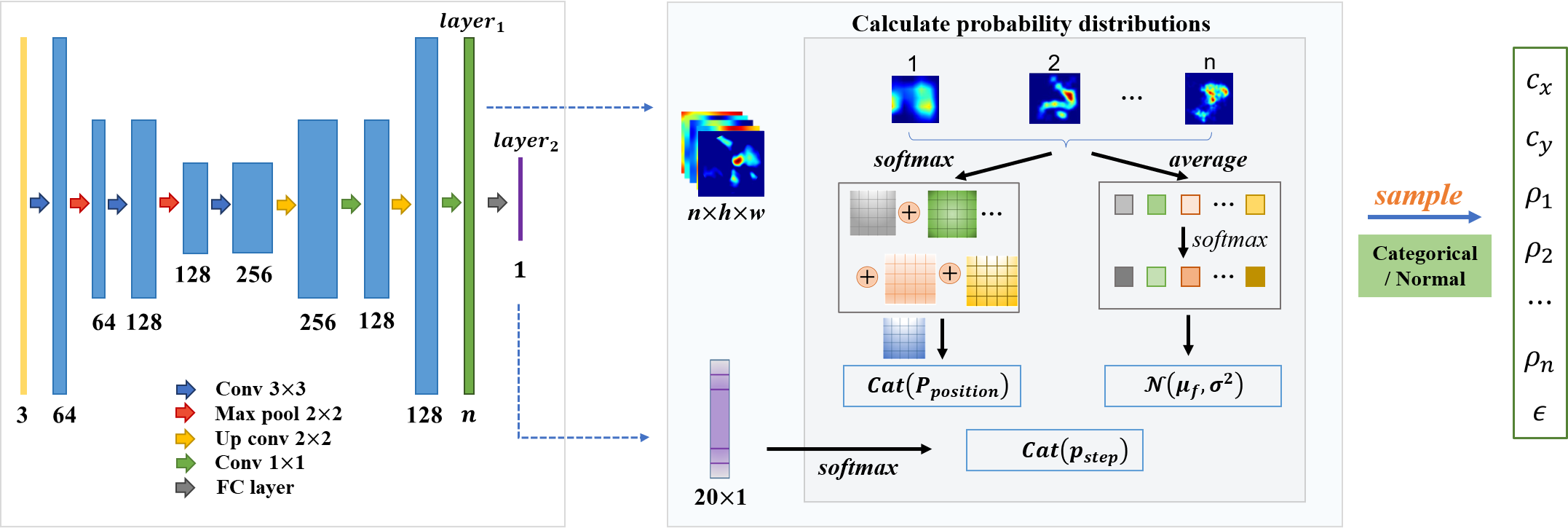}
\caption{ {The  details of the designed agent. The left shows the network structure of the agent. Based on the output feature maps $layer_{1}$ and $layer_{2}$, the right illustrates how they can be exploited to construct the policy probability distribution for each attack variable.}}
\label{fig:agent2}
\end{figure*}

In the face recognition task, given a clean face image $\bm{x}$, the goal of the adversarial attack is to make the face recognition model predict a wrong identity for the perturbed face image $\bm{x}^{adv}$. Formally, the perturbed face with the adversarial patch can be formulated as Eq. (\ref{eq:xadv}), where $\odot$ is Hadamard product and $\bm{\tilde x}$ is the adversarial perturbation across the whole face image. $\mathcal{A}$ is a  {binary} mask matrix to constrain the shape and pasting position of the patch, where the value of the patch area is 1.
\begin{equation}\label{eq:xadv}
\bm{x}^{a d v}=(\textbf{1}-\mathcal{A}) \odot \bm{x}+\mathcal{A} \odot \bm{\tilde x}.
\end{equation}
The previous methods either optimize $\bm{\tilde x}$ with pre-fixed $\mathcal{A}$, or fix $\bm{\tilde x}$ to select the optimal $\mathcal{A}$. In our method, we optimize $\mathcal{A}$ and $\bm{\tilde x}$ simultaneously to further improve the attack performance. 

For the optimization of the mask matrix $\mathcal{A}$, we fix the shape and size of the patch region in $\mathcal{A}$, and change the patch's upper-left coordinates  $c=\left(c_{x}, c_{y}\right)$  to adjust the mask matrix. In order not to interfere with the liveness detection module, we limit the pasting position to the area that does not cover the key facial features (i.e. eyes, eyebrows, nose, etc). Please refer to Section \ref{sec:opt} for details.

To generate the perturbation, the ensemble attack \cite{esm} based on MI-FGSM \cite{dong2018boosting} is   used here. 
For the ensemble attack with $n$ surrogate models, we let $\rho_{i}$ denote the weight of each surrogate model $f_{i}$ and $\epsilon$ denote the attack step size. Then taking the un-targeted attack (or dodging in the face recognition case) as an example, given the ground-truth identity $y$, we let $f_i(\bm{x},y)$ denote the confidence score that the model predicts a face image $\bm{x}$ as identity $y$, then $\bm{\tilde x}$ can be computed by an iteration way. Let $t$ denote the $t$-th iteration, then:
\begin{equation}\label{eq:mi_x}
\bm{x}^{adv}=\bm{\tilde x}_{t+1}=(1-\mathcal{A}) \odot \bm{x}+\mathcal{A} \odot \left (\bm{\tilde x}_{t}+\epsilon \cdot \bm{\emph{sign}}\left(g_{t+1}\right)\right),
\end{equation}
\begin{equation}\label{eq:mi_g}
g_{t+1}=\mu \cdot g_{t}+\frac{\nabla_{x} L\left(\bm{\tilde x}_{t}, y\right)}{\left\|\nabla_{x} L\left(\bm{\tilde x}_{t}, y\right)\right\|_{1}},
\end{equation}
\begin{equation}\label{eq:L1}
L\left(\bm{\tilde x}_t, y\right)=\sum_{i=1}^{n} \rho_{i} \cdot \ell\left(\bm{\tilde x}_t, y, f_{i}\right),
\end{equation}
where $\ell\left(\bm{\tilde x}_t, y, f_{i}\right)=1-f_{i}\left(\bm{\tilde x}_t, y\right)$ and $\sum_{i=1}^{n} \rho_{i}=1$.  For the targeted attack (or impersonation in the face recognition case), given the target identity $\widehat{y}$, $\ell$ can be simply replaced by $f_{i}\left(\bm{\tilde x}_t, \widehat{y}\right)$.

Our attack goal is to simultaneously optimize both the patch position and the perturbation to generate good adversarial patches to attack the target model. Therefore, the patch's coordinates  $c=\left(c_{x}, c_{y}\right)$, the attack step size $\epsilon$ in Eq. (\ref{eq:mi_x}) and weights $\rho_{i}$ in Eq. (\ref{eq:L1}) are set as the learned  {variables. We call them attack variables in the following sections.}
To be suitable for the target model, we adjust the attack variables dynamically through a small number of queries to the target model. The details of solving for these variables are shown in Section. \ref{sec:rl}.

\subsection{Attacks based on RL}\label{sec:rl}

\subsubsection{Formulation overview using RL}\label{seq:form}
 {In our method, the process to optimize the attack variables is formulated into the process of learning the agent's parameters in the RL framework under the guidance of the reward signal from the target model}. 
Specifically, the $t$-th variable' value is defined as the action $a_{t}$  {in the action space $a$} generated by the agent under the guidance of the policy $\pi$. The image input to the agent is defined as the state $s$, and the environment is the  {target} model $F(\cdot)$. $\pi_{\theta}(a \!\mid\! s)$ with parameters $\theta$  {is the policy function whose output is the probability value corresponding to each action $a_{t}$ in the action space $a$ under the state $s$. It obeys a certain probability distribution and is a rule used by the agent to decide what action to take.}

The reward reflects the performance of the currently generated adversarial patch on the target model, and the training goal of the agent is to learn good policies to maximize the reward signal. In the face recognition, the goal of the dodging attack is to generate images that are as far away as possible from the ground-truth identity $y$, while impersonation attacks want to generate images that are as similar as possible to the target identity $\hat{y}$. Thus, the reward function $R$ is formalized as:
\begin{equation}\label{eq:reward}
R=\left\{\begin{array}{ll}
\ell\left(\bm{x}^{a d v}, y, F\right)=1-F\left(\bm{x}^{a d v}, y\right) & \text { if dodging } \\
\ell\left(\bm{x}^{a d v}, \hat{y}, F\right)=F\left(\bm{x}^{a d v}, \hat{y}\right) & \text { if impersonation }
\end{array}\right.
\end{equation}
In the iterative training, the agent firstly predicts a set of  {actions} according to policy $\pi$, and then the adversarial patch based on the predicted  {actions} is generated. Finally the generated adversarial face image is input to the  {target} model to obtain the reward value. In this process, policy gradient  \cite{pol_grad} is used to guide the agent's  update.
After multiple training iterations, the agent will generate actions that perform well on the target model with a high probability.

\subsubsection{Design of the agent}\label{seq:agent}
The agent needs to learn the policies of the position, weights and attack step size. To take advantage of the  coupling relationship between the position and perturbation, the two factors need to be mapped to the feature map output by the \emph{same} agent at the same time.
Considering this, we design a U-net \cite{unet} based structure, which can achieve the correspondence between positions and image pixels, and output the feature map of the same size as the input image. Let the number of surrogate models be $n$, we design the agent to output the feature map with $n$ channels and the same length $h$ and width $w$ as the input image (i.e. the size is $n \times h \times w$). 

In each channel $M_{i}(i=1, \ldots, n)$ of the feature map $M$, the relative value of each pixel point represents the importance of each position for the surrogate model $f_{i}$, and the  {average value of the overall channel} reflects the importance of the corresponding surrogate model. We believe that the patch requires different attack strengths in different locations, so at the top layer of the agent network, a fully connected layer is used to map the feature map $M$ to a vector $V$ representing different attack step values. The structural details of the agent are shown in Figure \ref{fig:agent2}.

Specifically, for the position action, the optional range of positions is discrete, so the position policy $\pi_{\theta}^{1}$ is designed to follow Categorical distribution \cite{murphy2012machine}. Given the probability $P_{position}$ of each selected position, the position parameters 
$\left(c_{x}, c_{y}\right) \sim \bm{\emph{Cat}}\left(P_{\text {position }}\right)$, and $P_{position}$ is computed:
\begin{equation}\label{eq:p_pos}
P_{{position}}=\frac{1}{n} \sum_{i=1}^{n} \  \bm{\emph{softmax}}\left(M_{i}\right).
\end{equation}
For the weight actions, the weight ratio of the loss on each surrogate model $f_{i}$ to the ensemble loss is a continuous value, and we set the weight policy $\pi_{\theta}^{2}$ to follow Gaussian distribution \cite{murphy2012machine}. So the $i$-th weight parameter $\rho_{i} \sim \mathcal{N}\left(\mu_{f_{i}}, \sigma^{2}\right)(i\!=\!1, \ldots, n)$, and $\mu_{f_{i}}$ is calculated as:
\begin{equation}\label{eq:normal}
\mu_{f_{i}}=\bm{\emph{softmax}}\left(\overline{M_{1}}, \overline{M_{2}}, \ldots, \overline{M_{n}}\right)_{i},
\end{equation}
where $\overline{M_{i}}$ refers to the mean value of the $i$-th channel in the feature map, and $\sigma$ is a hyperparameter. In the actual sampling, we use the clipping operation to make the sum of weights equal to 1.

For the attack step action, we set 20 values in the range of 0.01 to 0.2 at intervals of 0.01, and adopt Categorical distribution \cite{murphy2012machine} as the step size policy $\pi_{\theta}^{3}$ due to the discreteness of the values. So the step size parameter $\epsilon \sim \bm{\emph{Cat}}\left(p_{step}\right)$, and probability $p_{step}$ of each candidate value is:
\begin{equation}\label{eq:step}
p_{step}=\bm{\emph{softmax}}\left(\bm{\emph{FC}}\left(P_{position}\right)\right).
\end{equation}
By sampling from the corresponding distribution, we can obtain $(c_x,c_y), \rho_{i}(i=1,...,n)$ and $\epsilon$. 

\subsubsection{Policy Update}\label{seq:pol_grad}

In the agent training, the goal is to make the agent $h_{\theta}$ learn a good policy  {$\pi_\theta=\left\{\pi_\theta^i \mid i=1, \ldots, T\right\}$} with parameters $\theta$ to maximize the expectation of the reward $R$,  {where $T$ represents the number of attack variables, and $T=3$ in our case (the attack variables are patch position $\left(c_{x}, c_{y}\right)$, attack step size $\epsilon$ in Eq. (\ref{eq:mi_x}) and weights $\rho_{i}$ in Eq. (\ref{eq:L1}), respectively.)} $\tau=\left(s, a_{1}, a_{2}, \ldots, a_{T}\right)$ is  {a set of decision results obtained by sampling according to the policy $\pi_{\theta}$ of the current agent, and is formalized as a set of the input state s and the sampling results of each attack variable}, then the optimal policy parameters $\theta^{*}$ can be formulated as:
\begin{equation}\label{eq:theta}
\theta^{*}=\arg \max _{\theta} J(\theta)=\arg\max _{\theta}\ E_{\tau \sim \pi_{\theta}(a \mid s)}[R(\tau)].
\end{equation}
We use the policy gradient \cite{pol_grad} method to solve for $\theta^{*}$ by the gradient ascent method, and follow the REINFORCE algorithm \cite{reinforce}, using the average value of $N$ sampling of the policy function distribution to approximate the policy gradient $\nabla_{\theta} J(\theta)$:
\begin{equation}\label{eq:pol_grad}
\begin{aligned}
\nabla_{\theta} J(\theta) &= \  {E_{\tau \sim \pi_{\theta}(a \mid s)}\left[\sum_{t=1}^{T} \nabla_{\theta} \log \pi_{\theta}^{t}\left(a_{t} \mid s\right) R(\tau)\right]} \\
&  {\approx \frac{1}{N} \sum_{n=1}^{N} \sum_{t=1}^{T} \nabla_{\theta} \log \pi_{\theta}^{t}\left(a_{t} \mid s\right) R_{n}},
\end{aligned}
\end{equation}
where $R_{n}$ is the reward in the $n$-th sampling.  {
A large reward will cause the agent parameters $\theta$ to have a large update along the current direction, while a small reward means that the current direction is not ideal, and the corresponding update amplitude will also be small.
}
Therefore, the agent can learn good policy functions with the update of $\theta$ in the direction of increasing the reward.

For actions that follow the Categorical policy  {(i.e., $\pi_{\theta}^{1}$ and $\pi_{\theta}^{3}$), let $p(a)$ denote the probability of action $a$ under the corresponding Categorical probability distribution (i.e., the probability value $p(a_{1})$ of the position variable under $\bm{\emph{Cat}}\left(P_{\text {position }}\right)$, and the probability value $p(a_{3})$ of the attack step size variable under $\bm{\emph{Cat}}\left(p_{step}\right)$)}, then for $\pi_{\theta}^{1}$ and $\pi_{\theta}^{3}$,  {$\nabla_{\theta} \log \pi_{\theta}^{t}(a_{t} \!\mid\! s)$} in Eq. (\ref{eq:pol_grad}) can be calculated as:
\begin{equation}\label{eq:pg_cat}
 {\nabla_{\theta} \log \pi_{\theta}^{t}(a_{t} \mid s)=\frac{d \  \log p(a_{t})}{d \theta}, \quad(t=1,3)}.
\end{equation}

For actions that follow the Gaussian policy distribution ( {i.e. the weights of the surrogate models that follow $\pi_{\theta}^{2}$}), the mean value $\mu_{f}$ of Gaussian distribution is calculated by the output of the agent, so $\mu_{f}$ can be expressed as $h_{\theta}(s)=\mu_{f}$. Therefore, for $\pi_{\theta}^{2}$ following the Gaussian policy,  {$\nabla_{\theta} \log \pi_{\theta}^{t}(a_{t} \!\mid\! s)$} in Eq. (\ref{eq:pol_grad}) can be calculated as follows:
\begin{equation}\label{eq:pg_normal}
\begin{aligned}
 {\nabla_{\theta} \log \pi_{\theta}^{t}(a_{t} \mid s)}
&=\nabla_{\theta}\left[-\frac{\left( {a_{t}}-h_{\theta}(s)\right)^{2}}{2 \sigma^{2}}-\log (\sigma)-\log (\sqrt{2 \pi})\right] \\
&=\frac{ {a_{t}}-h_{\theta}(s)}{\sigma^{2}} \cdot \frac{d h}{d \theta},  {\quad(t=2)}.
\end{aligned}
\end{equation}

After the policy update using Eq. (\ref{eq:pol_grad}), we solve for the optimal parameters $\theta^{*}$. And then the optimal action strategy is obtained. 

\subsection{Optional area of the pasting positions}\label{sec:opt}
In this section, we give more details about changing the pasting position of the adversarial patch. 

It is worth noting that in order not to interfere with the liveness detection module and to maintain the concealment of the attack method, the area that does not cover the facial features of the face (e.g. cheek and forehead) is regarded as the optional pasting area. In practical applications, the liveness detection module is often used in combination with the face recognition to confirm the real physiological characteristic of the object and exclude the form of replacing real faces with photos, masks, etc. \cite{DBLP:journals/corr/abs-2010-04145}. It is mainly based on the depth or texture characteristics of the face skin or the movements of the object (such as blinking and opening the mouth), so the patch cannot be pasted in the area that covers the facial features (such as eyes and mouth).

\begin{figure*}[h]
\centering\includegraphics[width=0.75\textwidth]{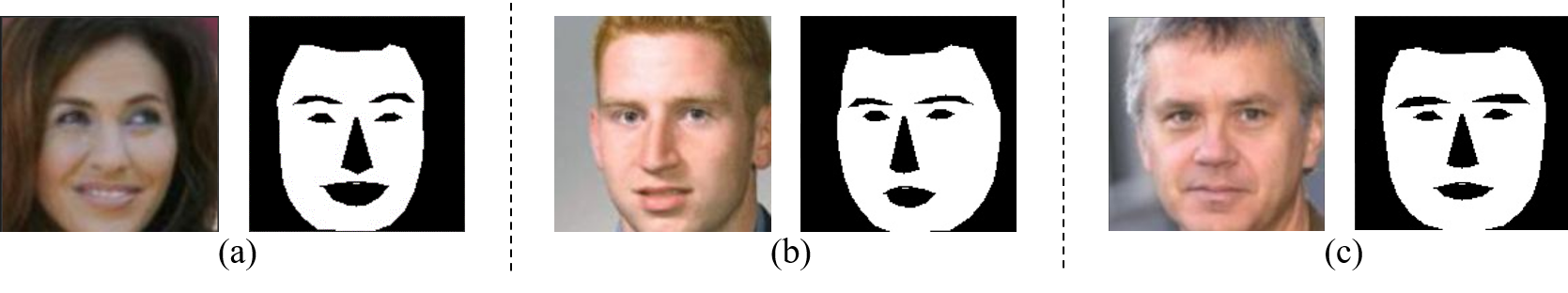}
\caption{Several groups of faces and the corresponding effective pasting areas. For each group, the left is the face image, and the white part of the right image represents the effective area.}
\label{fig:mask}
\end{figure*}

Specifically, we use ${\small{\tt dlib}}$ library to extract 81 face feature points  and determine the effective pasting region. Figure \ref{fig:mask} shows some examples of the effective pasting area corresponding to the face. 
After calculating the probability of each position $P_{position}$ in Eq. 
 (\ref{eq:p_pos}), we set the probabilities of the invalid positions to 0, and then sample the pasting position.

\subsection{Overall framework}
The complete process of our method is given in Algorithm \ref{alg:1}. In the $K$ iterations of the agent  {learning}, policy functions 
are firstly calculated according to the output of the agent, and then $N$ samplings are performed according to the probability distribution of the policy function to generate $N$ sets of parameters. According to each set of parameters, the attacks are conducted on surrogate models and the generated  adversarial examples are input to the face recognition model to obtain the reward.  {Policy functions are finally updated according to the rewards.} During this process, if a successful attack is achieved, the iteration is stopped early.

\begin{table*}[t]
\caption{Results of dodging (untargeted) attack and impersonation (targeted) attack against FaceNet, ArcFace34, ArcFace50, CosFace50, and commercial API in the black-box setting. We report ASR and average NQ required for simultaneous optimization. Three single models and their ensemble version are used as surrogate models (Note: for each column, the corresponding target model is not within the ensemble version).}\label{tab:fig1}
\renewcommand\arraystretch{0.9}
\centering 
\resizebox{0.85\linewidth}{!}{
\begin{tabular}{cc|cc|cc|cc|cc|cc}
\toprule[1.2pt]
\multicolumn{2}{l|}{\multirow{2}{*}{\diagbox{surrogate}{target}}}        & \multicolumn{2}{c|}{FaceNet} & \multicolumn{2}{c|}{ArcFace34} & \multicolumn{2}{c|}{ArcFace50} & \multicolumn{2}{c|}{CosFace50} & \multicolumn{2}{c}{cml. API} \\ \cline{3-12} 
\multicolumn{2}{c|}{}                              & {\small ASR}    & {\small \ NQ\ }        & {\small ASR}             & {\small \ NQ\ }   & {\small ASR}      & {\small \ NQ\ }      & {\small ASR}        & {\small \ NQ\ }   & {\small ASR}        & {\small \ NQ\ }        \\ \hline
\multicolumn{1}{c|}{\multirow{5}{*}{\rotatebox{90}{\small Dodging}}} & FaceNet   & \multicolumn{2}{c|}{-}       & 13.21\%         & 79         & 16.89\%         & 80         & 11.47\%                & 72   & 10.37\%                & 69             \\  
\multicolumn{1}{c|}{}                               & ArcFace34 & 84.97\%        & 17         & \multicolumn{2}{c|}{-}         & 57.28\%         & 18         &23.61\%                 & 52        & 32.31\%               & 53     \\ 
\multicolumn{1}{c|}{}                               & ArcFace50 & 89.37\%        & 14         & 61.05\%         & 16         & \multicolumn{2}{c|}{-}         & 33.45\%                & 45         & 38.20\%                & 49    \\ 
\multicolumn{1}{c|}{}                               & CosFace50 & 85.91\%        & 15         & 53.48\%         & 15         & 60.51\%         & 23         & \multicolumn{2}{c|}{-}       & 29.41\%                & 57  \\ 
\multicolumn{1}{c|}{}                               & Ensemble  & \textbf{96.65\%}        & 11         & \textbf{72.86\%}         & 18        & \textbf{72.09\%}         & 27         & \textbf{62.50\%}         & 23      & \textbf{52.19\%}                & 46   \\ \midrule[1pt]

\multicolumn{1}{c|}{\multirow{5}{*}{\rotatebox{90}{\small Impersonation}}}       & FaceNet   & \multicolumn{2}{c|}{-}       & 12.77\%         & 23         & 15.09\%         & 15         &8.70\%                 & 75            & 9.52\%                & 161 \\  
\multicolumn{1}{c|}{}       & ArcFace34 & 58.53\%        & 25         & \multicolumn{2}{c|}{-}         & 37.73\%         & 47         & 28.31\%       & 69        & 20.55\%                & 126   \\  
\multicolumn{1}{c|}{}                               & ArcFace50 & 53.52\%        & 28        & 33.23\%         & 60         & \multicolumn{2}{c|}{-}         & 28.30\%        &  53        & 27.91\%                & 114   \\ 
\multicolumn{1}{c|}{}                               & CosFace50 & 59.21\%        & 31         & 43.40\%         & 55         & 39.62\%         & 47         & \multicolumn{2}{c|}{-}       & 16.13\%                & 157  \\  
\multicolumn{1}{c|}{}                               & Ensemble  & \textbf{72.83\%}        & 27        & \textbf{50.28\%}         & 66         & \textbf{49.50\%}         & 36         & \textbf{40.08\%}        & 77      & \textbf{37.56\%}                & 91  \\ \bottomrule[1.2pt]
\end{tabular}}
\end{table*}

\begin{figure*}[t]
\centering
\includegraphics[width=0.85\textwidth]{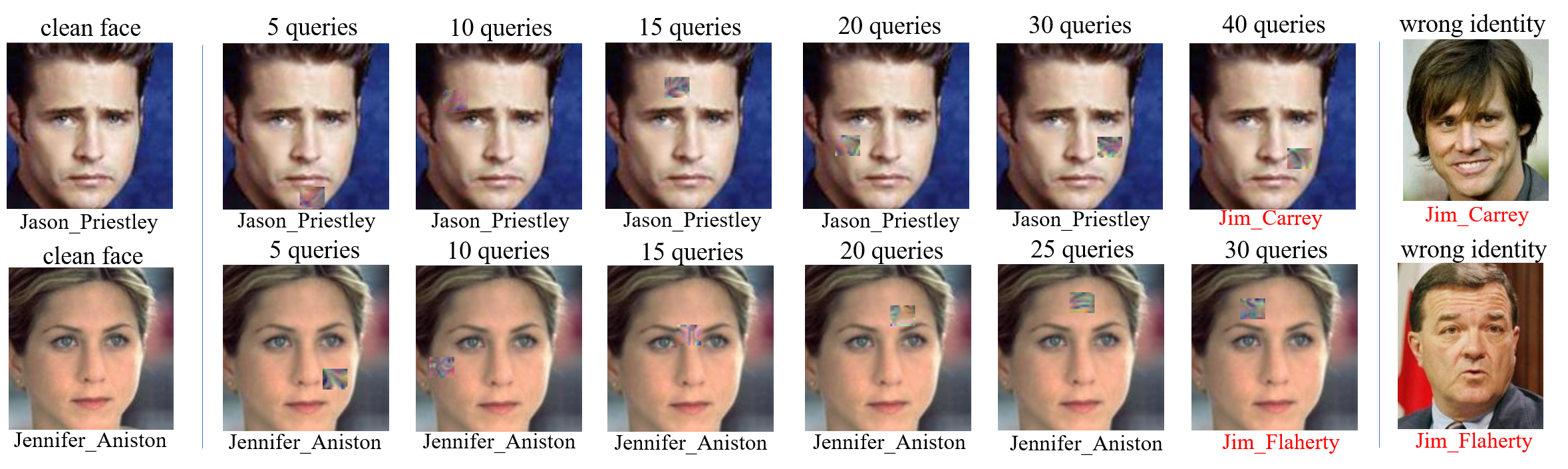}
\caption{Examples at different stages of the simultaneous optimization process. The black text at the bottom of images denotes the ground-truth identity, and the red text is the false identity after attacks.}
\label{fig:vis}
\end{figure*}

\begin{algorithm}[t]
	\renewcommand{\algorithmicrequire}{\textbf{Input:}}
	\renewcommand{\algorithmicensure}{\textbf{Output:}}
	\caption{Simultaneous optimization for the position and perturbation of the adversarial patch}
	\label{alg:1}
	\begin{algorithmic}[1]
		\REQUIRE {Face image $\bm{x}$, target $\hat{y}$, agent $h_{\theta}$, target model $F(\cdot)$, $n$ surrogate models $f_{i} \small(i\!=\!1, \ldots, n)$, iterations {\footnotesize$Z$} in {\small MI-FGSM}, sample times $N$, patch size $s_{h}, s_{w}$, learning rate $\alpha$, training epoch $K$.}
		\ENSURE $\bm{x}^{adv}$
		\FOR{$k$ = 1  to $K$}
    		\STATE { feature map $M \leftarrow h_{\theta}$; \ \ policies $\pi_{\theta}^{1}$, $\pi_{\theta}^{2}$, $\pi_{\theta}^{3}$ $\leftarrow$ according to Eq. (\ref{eq:p_pos}), Eq. (\ref{eq:normal}), Eq. (\ref{eq:step}) ;}
    		\FOR{$i$ = 1  to $N$}
        		\STATE {actions $\left(c_{x}, c_{y}, \rho_{1}, \rho_{2}, \ldots, \rho_{n}, \epsilon\right) \leftarrow $ sampling from $\pi_{\theta}^{1}$, $\pi_{\theta}^{2}$, $\pi_{\theta}^{3}$;\ \
        		$\mathcal A \leftarrow $ $(c_{x}, c_{y})$ and $s_{h}, s_{w}$ ;}
        		\FOR{$t$ = 1  to $Z$}
                    \STATE {$g_{t}$ $\leftarrow$ according to Eq. (\ref{eq:L1}) and Eq. (\ref{eq:mi_g}) ; $\bm{\tilde x}_t$ $\leftarrow$  according to Eq. (\ref{eq:mi_x});}
        		\ENDFOR
        		\STATE {Update $R_{i}$ $\leftarrow$ according to Eq.  (\ref{eq:reward});}
    		\ENDFOR
    		\STATE {$\small \nabla_{\theta} J(\theta)$ \!$\leftarrow$\! {\small according to} Eq. (\ref{eq:pol_grad});
    		$\small \theta$ \!$\leftarrow$\! $\small \theta+\alpha \cdot \nabla_{\theta} J(\theta)$;}
    		\STATE \textbf{if} \ $F\left(\bm{x}^{a d v}\right)\!=\!\hat{y}$ \ \ \textbf{then} \ \  break;
		\ENDFOR
		\STATE {\textbf{return} $\bm{x}^{adv}$}
	\end{algorithmic}  
\end{algorithm}

\section{Combined with gradient estimation attack}\label{sec:extend}
Our simultaneous optimization method can also be combined with the gradient estimation  {attacks} (e.g., Zeroth-Order (ZO) optimization \cite{zo-adamm,chen2017zoo}, natural evolution strategy \cite{nes}, random gradient estimation \cite{random})  {to further enhance the attack performance. Gradient estimation  {attack} estimates the gradient of each pixel through the information obtained by querying the model. It has good attack performance, but the query cost is high.
The result of the simultaneous optimization can be used as the initialization value of the gradient estimation to provide a good initial position and pattern. Compared with the random initialization used in the original gradient estimation, a good initialization value with certain attack performance can improve the query efficiency.} Here we take Zeroth-Order (ZO) optimization \cite{zo-adamm,chen2017zoo} as an example to describe the combined manner.

In the Zeroth-Order (ZO) optimization,  {let $\rm{\bm{x}}$ denote the image variable, and its value is the adversarial image in the iterative process. The value range in $\rm{\bm{x}}$ is $[0,1]$. To expand the optimization range, $\rm{\bm{x}}$ is often replaced \cite{chen2017zoo} as:
\begin{equation}\label{eq:g_w}
\rm\bm{x}=\frac{1}{2}(1+\tanh \bm{\phi}).
\end{equation}
In this way, optimizing $\rm{\bm{x}}$ turns to optimize $\bm{\phi}$, and the optimization range is expanded to $[-\infty,+\infty]$.
Gradient estimation is achieved by adding a small offset at $\bm{\phi}$ for the calculation of the symmetric difference quotient \cite{lax2014calculus,chen2017zoo}.
In order for the addition of the small offset to have an effect on the result, $\bm{\phi}$ cannot be in an overly smooth position in the solution space, i.e.,
}
the gradient of $\rm{\bm{x}}$ at $\bm{\phi}$ (i.e. $\nabla_{\bm{\phi}} \rm{\bm{x}}$) can not be too small. $\nabla_{\bm{\phi}} \rm{\bm{x}}$ is calculated as follows:
\begin{equation}\label{eq:g_w}
\nabla_{\bm{\phi}} \rm{\bm{x}}=\frac{1}{2}\left(1-\tanh ^{2}(\bm{\phi})\right).
\end{equation}
 {
To make the initial values of the gradient estimation obtained by simultaneous optimization also meet the above requirements, we need to add this gradient requirement to the objective function of the simultaneous optimization.}
So the loss function $L(\cdot, \cdot)$ in Eq. (\ref{eq:L1}) is modified as:

\begin{equation}\label{eq:L}
L\left(\bm{\tilde x}_{t}, y\right)=\sum_{i=1}^{n} \rho_{i} \cdot \ell\left(\bm{\tilde x}_{t}, y, f_{i}\right)+\frac{\beta}{s_{h} \cdot s_{w}} \sum_{i=1}^{s_{h}} \sum_{j=1}^{s_{w}} \nabla_{\phi_{\langle i j\rangle}} \rm{\bm{x}},
\end{equation}
where $s_{h}$ and $s_{w}$ represent the height and width of the perturbation patch pasted to the face, and $\beta$ is the scale factor. Given the paste coordinates $\left(c_{x}, c_{y}\right)$,  {then at the pixel point $(i,j)$ of the patch}, $\phi_{\langle i j\rangle}={\rm arctanh}\left(2 \cdot \bm{\tilde x}_{t\left(c_{y}+i, c_{x}+j\right)}-1\right)$.

 {Replacing Eq. (\ref{eq:L1}) with Eq. (\ref{eq:L}), keeping the rest of the process in Algorithm 1 unchanged, we can obtain an adversarial initialization result (i.e. pattern $\bm{\tilde x}^{*}$ and position $\left(c_{x}^{*}, c_{y}^{*}\right)$) suitable for gradient estimation through our simultaneous optimization.
On this basis, we fix the position $\left(c_{x}^{*}, c_{y}^{*}\right)$ and use the gradient estimation method to only refine the perturbation. Through querying the target models, we can estimate the gradient $\hat{g}_{t}$ in $t$-$th$ iteration:
\begin{equation}\label{eq:zoo2}
\hat{g}_{t}=\frac{\ell\left( \bm{\phi}_{t}+\varepsilon \bm{e}_{t}\right)-\ell\left(\bm{\phi}_{t}-\bm{\varepsilon} \bm{e}_{t}\right)}{2 \varepsilon},
\end{equation}}
 {where $\varepsilon$ is a small constant and $\bm{e}_{t}$ is a standard basis vector where pixels selected for update correspond to the value of 1.
$\ell\left( \bm{\phi}_{t} \pm \varepsilon \bm{e}_{t}\right)$ is the loss function, that is, the variant of $\ell\left(\bm{\tilde x}_t, y, f_{i}\right)$ of Eq. (\ref{eq:L1}) migrated from $x$-space to $\phi$-space, and the result is obtained by querying the model.
$\phi_{0}={\rm arctanh}\left(2 \cdot \bm{\tilde x}^{*}-1\right)$.
Based on $\hat{g}_{t}$, we can use ADAM optimization method \cite{chen2017zoo} to optimize image pixel values in the iterative process.}  {Thanks to the good initialization value, the query efficiency in the gradient estimation attack will also be improved.}

\section{Experiments}\label{sec:exp}
In this section, we give the experiments from experimental settings,  comparisons with SOTA methods, ablation study, integration with gradient estimation attacks, and so on. 

\subsection{Experimental Settings}\label{sec:expset}

\textbf{Target models and datasets:} We choose five face recognition models as target models, including four representative open-source face recognition models (i.e. FaceNet \cite{facenet}, CosFace50 \cite{cosface}, ArcFace34 and ArcFace50 \cite{arcface}) and one commercial face recognition API\footnote{https://intl.cloud.tencent.com/product/facerecognition}.
When performing ensemble attacks for the API, the four open-source models are set as surrogate models. \textbf{When performing ensemble attacks for one of these four models, the other three models are used as surrogate models}. Thus, the attacks are conducted via the transferability. 
Our face recognition task is based on the face identification mechanism, therefore, a large face database is needed. So, we randomly select 5,752 different people from  Labeled Faces in the Wild (LFW)\footnote{http://vis-www.cs.umass.edu/lfw/} and CelebFaces Attribute  (CelebA)\footnote{http://mmlab.ie.cuhk.edu.hk/projects/CelebA.html} to construct the face database. We use the above face models to extract the face features, and then calculate the cosine similarity with all identities in the face database to perform the 1-to-N identification. 
The identity with the highest similarity is regarded as the face's identity, and the corresponding similarity is taken as the confidence score.

\textbf{Metrics:} Two metrics, attack success rate (ASR) and the number of queries (NQ) are used to evaluate the performance of attacks. ASR refers to the proportion of images that are successfully attacked in all test face images, where we ensure that the clean test images selected in the experiment can be correctly identified.
We count the cases of incorrect identification as successful attacks.
NQ refers to the number of queries to the target model required by the adversarial patch that can achieve a successful attack.

\textbf{Implementation:}

The size of the adversarial patch is set as $s_{h}=25, s_{w}=30$, the number of sampling $\small N$ in the policy gradient method is set to $5$, and the variance $\sigma$ of the Gaussian policy is equal to $0.01$. Other parameters are set as ${\small Z=150}$, $\alpha=0.001$, and ${\small K=50}$.

\subsection{Experimental results}
\subsubsection{Performance of simultaneous optimization}
We first evaluate our simultaneous optimization method qualitatively and quantitatively according to the setting in Section \ref{sec:expset}.
Table \ref{tab:fig1} shows the quantitative results of dodging and impersonation attacks under the black-box setting against the five target models. 

\begin{table*}[ht]
\caption{Comparison results of the ASR and NQ between our method and other adversarial patch methods that only change perturbations through transferability (GenAP) or gradient estimation (ZO-AdaMM) with fixed positions, only change the position (RHDE) with fixed perturbations,  change position and perturbations alternately (LO), and two-step manner (PatchAttack).}\label{tab:sota}
\renewcommand\arraystretch{1.0}
\centering 
\begin{threeparttable}
\resizebox{0.95\linewidth}{!}{
\begin{tabular}{l|cc|cc|cc|cc|cc|cc}
\toprule[1.2pt]
\multirow{3}{*}{\diagbox{method}{target}}         & \multicolumn{6}{c|}{Dodging}                & \multicolumn{6}{c}{Impersonation}          \\ \cline{2-13} 
         & \multicolumn{2}{c|}{FaceNet}  & \multicolumn{2}{c|}{ArcFace50} & \multicolumn{2}{c|}{CosFace50} & 
         \multicolumn{2}{c|}{FaceNet} & \multicolumn{2}{c|}{ArcFace50} & \multicolumn{2}{c}{CosFace50} \\ 
         & {\small ASR}    & {\small \ NQ\ }        & {\small ASR}             & {\small \ NQ\ }   & {\small ASR}      & {\small \ NQ\ }      & {\small ASR}        & {\small \ NQ\ }& {\small ASR}        & {\small \ NQ\ }& {\small ASR}        & {\small \ NQ\ }        \\ \hline
GenAP \cite{xiao2021improving} (\emph{p.})    & 58.50\% &  -   & 57.00\%  &  -  & 56.50\% &  -   & 28.75\%  &  -  & 27.50\%  &  -   & 24.25\%  &  -   \\ 
ZO-AdaMM \cite{zo-adamm} (\emph{p.}) & 65.79\% &1972  & 41.58\%  &2161 & 43.47\% &2434  & 50.52\% &2874 & 40.41\% &3198  & 39.16\% &3106 \\ 
RHDE \cite{advsticker} (\emph{l.})     & 48.82\% &408  & 35.39\%  &504 & 42.93\% &514  & 41.63\% &522 & 29.40\% &586  & 35.64\% &577  \\ 
LO \cite{rao2020adversarial} (\emph{p.l.})      & 76.82\% &1105 & 61.59\%  &1475 & 55.14\% &2117  & 55.69\% &1974 & 42.67\% &2870  & 38.66\% &2624 \\ 
PatchAttack \cite{yang2020patchattack} (\emph{p.l.})      &94.54\%     &1204     &65.44\%      &1566     &61.43\%      &1632     &68.95\%    &2389     &44.00\%    &3324       &35.56\%      &3878     \\ 
Ours  (\emph{p.l.})    & \textbf{96.65\%} &\textbf{11}  & \textbf{72.09\%}  &\textbf{27} & \textbf{62.50\%} &\textbf{23}  & \textbf{72.83\%} &\textbf{27} & \textbf{49.50\%} &\textbf{36}  & \textbf{40.08\%} &\textbf{77}  \\ 
\toprule[1.2pt]
\end{tabular}
}
\begin{tablenotes}
\item\footnotesize \quad \emph{l.}- change location. \ \emph{p.}- change perturbations 
\end{tablenotes}
\end{threeparttable}
\end{table*}

From the results in Table \ref{tab:fig1}, we can see : (1) the proposed method achieves good attack success rates and query efficiency under both two kinds of attacks. The dodging attack achieves the highest success rate of 96.65\% under 11 queries, while the impersonation attack achieves the highest success rate of 72.83\% under 27 queries. (2) The performance of using ensemble models to attack is better than that using a single model as the surrogate model, which shows that simultaneous optimization can adaptively adjust the weights of different surrogate models to achieve the best performance.
(3) For the relationship between the surrogate model and the target model,
when the two are structurally similar, it is more likely to help the attack. For example, for ArcFace50, CosFace50 has a greater influence on it. This is because the backbone of ArcFace50 and CosFace50 are both ResNet50-IR \cite{arcface}, and the backbone of FaceNet and ArcFace34 are Inception-ResNet-v1 \cite{inception} and ResNet34-IR \cite{arcface} respectively. 
Therefore, in ensemble attacks, we can dynamically adjust the importance of different models by optimizing their weights. 
(4) On the commercial API, performance drops slightly compared to the open-source model, but remains at an acceptable level. This is because commercial FR services introduce some defense measures like image compression.

\begin{figure*}[h]
\centering
\includegraphics[width=0.7\textwidth]{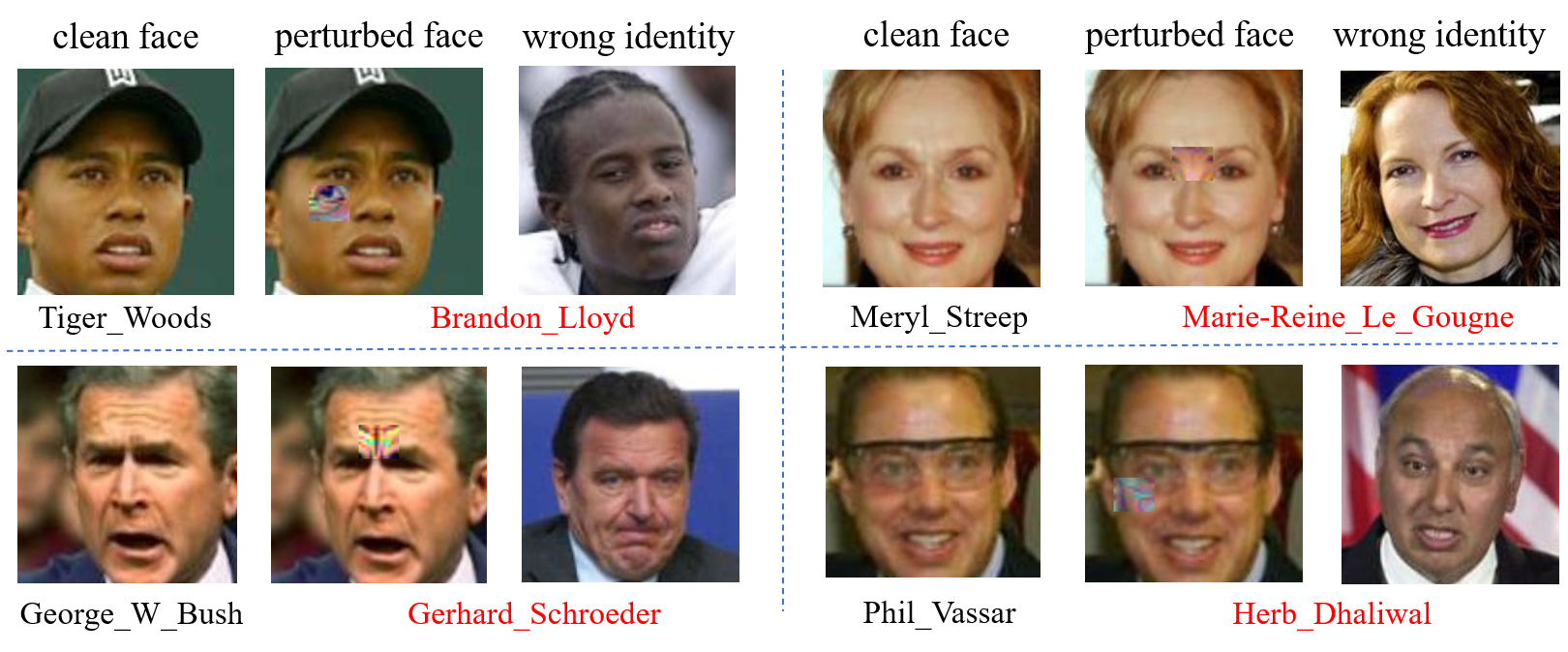}
\caption{ {More visual examples output by our method. The black text below the image denotes the ground-truth identity, and the red text is the false identity after attacks.}}
\label{fig:visa}
\end{figure*}

Figure \ref{fig:vis} shows some visual examples of the position and perturbation at different stages of the attack. We can see that with the increase of query times, the positions and perturbations of the generated adversarial patch are gradually stable and convergent until the face recognition model predicts the target wrong identity. This show that the proposed method has  a good convergence.  
More visual results are shown in Figure \ref{fig:visa}. For each group of three images, the first one represents the clean image, the second one represents the image after the attack, and the third one represents the image corresponding to the wrong identity in the face database. The above results are obtained by ensuring that the patch is not pasted to the area that covers the facial features. 

 {In addition, we also study the regularity of the position of the patch. We divide the face into four areas: between the eyebrows, the forehead, the left face and the right face, and count the position of the patch when attacking FaceNet, ArcFace50, and CosFace50. The ratio of patches in each area is shown in Table \ref{tab:rat}.
It can be seen that for the three models, the proportions of the patch located between the eyebrows are the highest, reaching 42.55\%, 58.32\%, 54.36\%, respectively, followed by the forehead, while the proportions of the left and the right face are relatively low.
This may be because  the position between the eyebrows is close to the key facial features (eyes, eyebrows), so these positions have a greater impact on the model's  prediction.
}
\begin{table}[t]
\caption{ {The ratio of patches in four areas (i.e. between the eyebrows, the forehead, the left face and the right face) when attacking FaceNet, ArcFace50, and CosFace50.}}
\centering  
\resizebox{0.9\linewidth}{!}{
\begin{tabular}{c|c|c|c}
\toprule[1.2pt]
                     & FaceNet & ArcFace50 & CosFace50 \\ \hline
Between the eyebrows & 42.55\% & 58.32\%   & 54.36\%   \\ \hline
Forehead             & 40.43\% & 25.01\%   & 26.73\%   \\ \hline
Left face            & 10.64\% & 5.56\%    & 10.94\%   \\ \hline
Right face           & 6.38\%  & 11.11\%   & 7.97\%    \\ \bottomrule[1.2pt]
\end{tabular}}\label{tab:rat}
\end{table}

\subsubsection{Comparisons with SOTA methods}
To prove the superiority, we compare our method with five state-of-the-art methods: GenAP \cite{xiao2021improving} and ZO-AdaMM \cite{zo-adamm} fix the position and change perturbations relying on transferability and gradient estimation, respectively. RHDE \cite{advsticker} fixes the perturbations and only changes the position, and Location Optimization (LO) \cite{rao2020adversarial} optimizes the position and the perturbation alternately, as well as the PatchAttack \cite{yang2020patchattack} optimizes the position and perturbation in a two-step manner. 
In the implementation, to adapt Location Optimization (LO) \cite{rao2020adversarial} to black-box attacks, we retain the framework of alternate iterations, and use gradient estimation to replace white-box gradient calculations.
 For GenAP, because we use its transferability to perform the black-box attack, its NQ is zero, and we use ``-" to replace it. For our method, the target model in each column is not within  the corresponding ensemble surrogate models (see Table \ref{tab:fig1}), therefore, the comparison is fair.  The above results on three target models are shown in Table \ref{tab:sota}.

\begin{figure}[t]
\centering\includegraphics[width=0.4\textwidth]{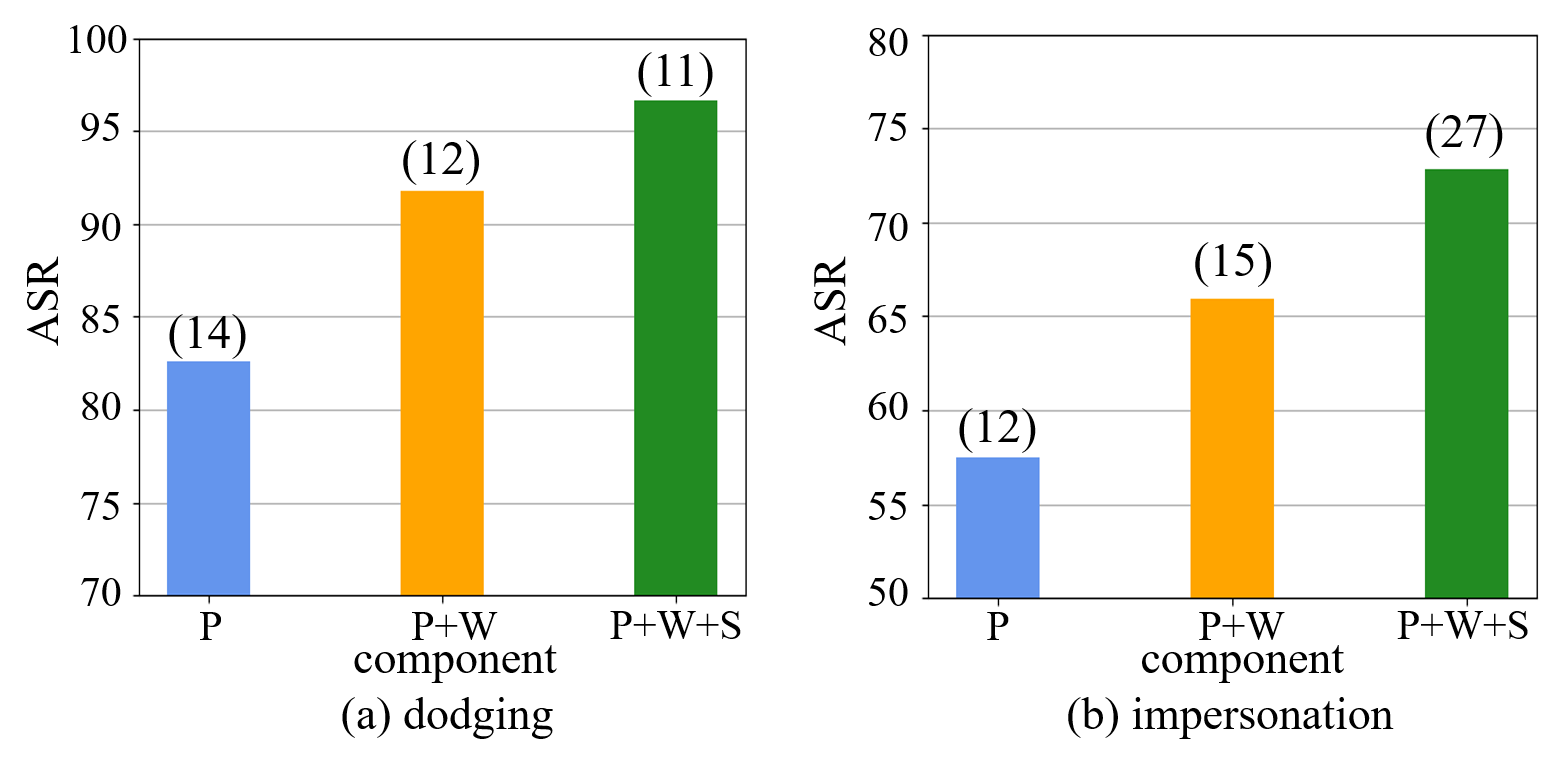}
\caption{ASR and the NQ (shown in brackets) when using Position (P), Position and weights (P+W), and Position, weights and step size (P+W+S) respectively.}
\label{fig:ablation}
\end{figure}


From the results, we can see that: (1) For the perturbation-only method, GenAP and ZO-AdaMM have an average success rate of 42.08\% and 46.82\%; For the location-only method, RHDE is 38.97\%. However, LO, PatchAttack, and ours are 55.10\%, 61.60\% and 65.61\%, respectively. Therefore, considering  both factors can achieve a significant improvement. (2) Our simultaneous optimization is  ahead of LO's alternate optimization and PatchAttack's two-step attack, which shows that simultaneous optimization can make better use of the inherent connection between the perturbation and location, so it is better than optimizing the two separately.
(3) Our method achieves the optimal query efficiency among several methods, requiring only a few dozen queries.

\subsubsection{Ablation study}\label{sec:ablation}

In order to verify the effectiveness of each attack variable used in our method, we also report the results when each component is added separately. 
First, we fix weights as equal values and the step size as 0.1 to test the performance when only changing the position parameter. Then, we add the weight parameter to learn the importance of each surrogate model. Finally, we add the step size parameters to carry out the overall learning.
The results in Figure \ref{fig:ablation} show that learning only the position parameter can achieve an average success rate of 69.80\%. The performance is greatly improved after adding the weight, and further improved after adding the step size. All parameters contribute to the enhancement of the overall attack effect and the weight parameter has a greater impact on the results than the step size. In the process of adding components, the number of queries (shown in brackets) is basically maintained at the same level.


\subsubsection{Performance against defense methods}\label{sec:defense}
 {Here, we test the  attacks' performance under two mainstream forms of defenses: preprocessing  and adversarial training.}

 {For the preprocessing defense, we explore the ``detecting-repairing" methods, i.e., they first detect the region of the adversarial patch, and then repair the patch via different strategies. Specifically,  we use three methods to defend our attack: Local Gradients Smoothing (LGS) \cite{lgs} locates the noisy region based on the concentrated high-frequency changes introduced during the attack, and regularizes the gradient of this region to suppress the effect of adversarial noise. DW \cite{dw} locates local perturbations by the dense clusters of saliency maps in adversarial images, and then blocks the corresponding regions for defense. Februus \cite{februus} locates the influential region based on the heat map of GradCAM, removes the region and conducts image restoration for defense.
The corresponding attack success rates (ASR) of dodging attacks against the above defense methods are shown in Table \ref{tab:defense1}.}

 {It can be seen that these methods cannot effectively weaken the attack performance.
For LGS and Februus, ASR only drops slightly, by 1.01\% and 3.37\% on FaceNet, respectively. For DW, ASR decreases by 14.94\% on FaceNet, but remains at an acceptable level with the ASR of 81.71\%.
We think this is because these methods do not locate/detect the local perturbations accurately.
In the face classification task, the local patches often make the model decision solely affected by this small area \cite{dw}, and thus localization is performed by the sensitive information exhibited by these local regions (e.g. frequency, saliency maps, and heatmaps of GradCAM).
However, in the face recognition task, the model's decision is affected by the combined effects of multiple regions on the face, and the addition of patches only affects the characteristics of a local facial feature. Before and after the attack, the region that the model focuses on is still scattered on each key area of the face instead of a certain area. After the attack, only the importance of a certain area will change slightly, so it is difficult to distinguish how it is different from the clean face image.
}

\begin{table}[t]
\caption{ {The attack success rate (ASR) of dodging attacks against three defense methods: LGS \cite{lgs}, DW \cite{dw}, and Februus \cite{februus}. The values in brackets show the change values compared to the undefended situation.}}
\centering  
\resizebox{0.68\linewidth}{!}{
\begin{tabular}{c|c|c|c}
\toprule[1.2pt]
                          & FaceNet               & ArcFace50             & CosFace50             \\ \hline
\multirow{2}{*}{LGS}     & 95.64\%               & 64.52\%               & 56.82\%               \\
                         & ($\downarrow$1.01\%)  & ($\downarrow$7.57\%)  & ($\downarrow$5.68\%)  \\ \hline
\multirow{2}{*}{DW}      & 81.71\%               & 46.10\%               & 45.45\%               \\
                         & ($\downarrow$14.94\%) & ($\downarrow$25.99\%) & ($\downarrow$17.05\%) \\ \hline
\multirow{2}{*}{Februus} & 93.28\%               & 66.22\%               & 46.88\%               \\
                         & ($\downarrow$3.37\%)  & ($\downarrow$5.87\%)  & ($\downarrow$15.62\%) \\  \bottomrule[1.2pt]
\end{tabular}}\label{tab:defense1}
\end{table}

\begin{table}[t]
\caption{ {Results of our method and GDPA \cite{xiang2021gdpa} under adversarial training. Results include the attack success rate (ASR) and the number of queries (NQ) of dodging attacks, and the change values compared to the undefended situation (shown in brackets). (GDPA is a white-box attack and does not involve NQ.)}}\label{tab:defense}
\centering  
\resizebox{0.95\linewidth}{!}{
\begin{tabular}{c|cc|cc|cc}
\toprule[1.2pt]
                      & \multicolumn{2}{c|}{FaceNet}                                    & \multicolumn{2}{c|}{ArcFace50}                                  & \multicolumn{2}{c}{CosFace50}                                   \\ \hline
                      & \multicolumn{1}{c|}{ASR}                   & NQ                 & \multicolumn{1}{c|}{ASR}                   & NQ                 & \multicolumn{1}{c|}{ASR}                   & NQ                 \\ \hline
\multirow{2}{*}{GDPA} & \multicolumn{1}{c|}{51.49\%}               & \multirow{2}{*}{-} & \multicolumn{1}{c|}{13.30\%}               & \multirow{2}{*}{-} & \multicolumn{1}{c|}{11.66\%}               & \multirow{2}{*}{-} \\
                      & \multicolumn{1}{c|}{($\downarrow$48.51\%)} &                    & \multicolumn{1}{c|}{($\downarrow$74.89\%)} &                    & \multicolumn{1}{c|}{($\downarrow$61.61\%)} &                    \\ \hline
\multirow{2}{*}{Ours} & \multicolumn{1}{c|}{93.96\%}               & 9                  & \multicolumn{1}{c|}{72.71\%}               & 20                 & \multicolumn{1}{c|}{59.72\%}               & 25                 \\
                      & \multicolumn{1}{c|}{($\downarrow$2.69\%)}  & ($\downarrow$ 2)   & \multicolumn{1}{c|}{($\uparrow$0.62\%)}    & ($\downarrow$ 7)   & \multicolumn{1}{c|}{($\downarrow$2.78\%)}  & ($\uparrow$ 2)    \\ \bottomrule[1.2pt]
\end{tabular}}
\end{table}

 {For adversarial training, we retrain the target model using adversarial examples generated by attack methods against this model, and then take the retrained robust model  as the new target model to test the attack performance \cite{rao2020adversarial}. To demonstrate our superiority, we compare our method with  GDPA \cite{xiang2021gdpa} under the same adversarial training setting. GDPA \cite{xiang2021gdpa} is a white-box attack method that also considers both positions and perturbations of the adversarial patch.  
The results of the attack success rate (ASR) and the number of queries (NQ) are shown in Table \ref{tab:defense}, where
the numbers in brackets denote the change value compared to the undefended model.}
 {It can be seen that adversarial training  defends against GDPA attack well, when we use the same GDPA to attack the robust target models, the ASRs have a great drop (48.51\% for FaceNet, 74.89\% for ArcFace50, and 61.61\% for CosFace50). This is reasonable because the generated adversarial examples by GDPA are integrated into the training set for these three models, and the target models have ``seen" the adversarial examples. As a contrast, adversarial training does not defend against our attack, the ASR only has a slight change (i.e. 2.69\%, 0.62\%, 2.78\% respectively), and it is still maintained at a relatively high level. The reason for this difference is that the patch's positions chosen by GDPA on face images usually lie in the key facial features like eye, mouth, nose, etc. In other words, the learned position is constrained within a fixed scope (its original paper also shows this phenomenon). While  the patch's position chosen by our method can occur in any face area except for the key facial features. Therefore, it is difficult for adversarial training to learn the complete data distribution prevalent in our attack, which leads to the poor defense performance. }

\begin{table}[t]
\caption{ {The attack success rate (ASR) of dodging attacks on FaceNet and ArcFace50 models when patches are translated or rotated.}}
\centering  
\resizebox{1\linewidth}{!}{
\begin{tabular}{c|ccc|cc}
\toprule[1.2pt]
                           & \multicolumn{3}{c|}{Translations}                                                                              & \multicolumn{2}{c}{Rotations}                                     \\ \cline{2-6} 
                           & \multicolumn{1}{c|}{2pix}                 & \multicolumn{1}{c|}{3pix}                  & 5pix                  & \multicolumn{1}{c|}{$5^{\circ}$}                    & $15^{\circ}$                    \\ \hline
\multirow{2}{*}{FaceNet}   & \multicolumn{1}{c|}{96.65\%}              & \multicolumn{1}{c|}{90.59\%}               & 79.24\%               & \multicolumn{1}{c|}{93.12\%}              & 83.21\%               \\
                           & \multicolumn{1}{c|}{($\downarrow$0.00\%)} & \multicolumn{1}{c|}{($\downarrow$5.49\%)}  & ($\downarrow$16.84\%) & \multicolumn{1}{c|}{($\downarrow$2.96\%)} & ($\downarrow$12.87\%) \\ \hline
\multirow{2}{*}{ArcFace50} & \multicolumn{1}{c|}{71.36\%}              & \multicolumn{1}{c|}{56.91\%}               & 53.11\%               & \multicolumn{1}{c|}{63.85\%}              & 53.20\%               \\
                           & \multicolumn{1}{c|}{($\downarrow$0.72\%)} & \multicolumn{1}{c|}{($\downarrow$15.17\%)} & ($\downarrow$18.97\%) & \multicolumn{1}{c|}{($\downarrow$8.23\%)} & ($\downarrow$19.46\%) \\ \bottomrule[1.2pt]
\end{tabular}}\label{tab:trans}
\end{table}

\begin{figure*}[t]
\centering
\includegraphics[width=1\textwidth]{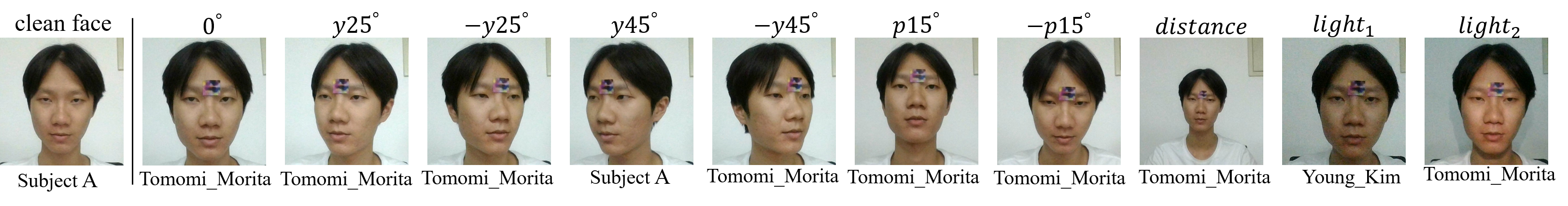}
\caption{Visual examples of the physical impersonation attack under different conditions. The text at the bottom of images denotes the recognition result of the FaceNet model.}
\label{fig:phy}
\end{figure*}

\begin{table*}[t]
\centering
\quad
\begin{minipage}[]{0.26\textwidth}
\caption{Success rate of dodging (D) and impersonation (I) attack for the frontal face in the physical world.}\label{tab:front}
\centering
\resizebox{\textwidth}{12mm}{
\begin{tabular}{c|c|c}
\toprule[1.2pt]
           & D        & I       \\ \midrule[1pt]
FaceNet    & 100.00\% & 83.33\% \\
ArcFace34  & 75.29\%  & 65.08\% \\
CosFace50  & 61.95\%  & 46.71\%        \\
cml. API  & 33.16\%  & 25.23\%  \\ \bottomrule[1.2pt]
\end{tabular}}
\end{minipage}
\qquad
\quad
\begin{minipage}[]{0.56\textwidth}
\caption{ {Success rate of dodging (D) and impersonation (I) attack in the physical world when changing distance, lighting and face postures including frontal ($0^{\circ}$), yaw angle rotation $\pm 25^{\circ}$, $\pm 45^{\circ}$, and pitch angle rotation $\pm 15^{\circ}$.}}\label{tab:env}
\centering
\resizebox{\textwidth}{12mm}{
\begin{tabular}{c|c|cccccc}
\toprule[1.2pt]
\multicolumn{2}{c|}{}          & $0^{\circ}$ & $\pm \ y 25^{\circ}$ & $\pm \ y45^{\circ}$ & $\pm \ p15^{\circ}$ & dist. & light       \\ \midrule[1pt]
\multirow{2}{*}{D} & FaceNet   & 100.00\% & 92.31\% & 36.36\% & 87.50\% & 97.06\% & 99.79\% \\
                   & ArcFace34 & 75.29\%  & 62.50\% & 28.57\% & 45.24\% & 71.43\% & 65.45\% \\ \hline
\multirow{2}{*}{I} & FaceNet   & 83.33\%  & 70.23\% & 35.75\% & 62.50\% & 88.24\% & 41.38\% \\
                   & ArcFace34 & 65.08\%  & 54.17\% & 23.81\% & 57.14\% & 60.71\% & 41.81\% \\ \bottomrule[1.2pt]
\end{tabular}}
\end{minipage}
\end{table*}

\begin{table*}[h]
\caption{The ASR and NQ of attacking FaceNet using adversarial patch output by simultaneous optimization (SO), simultaneous optimization (SO) combined with Zeroth-Order (ZO) optimization (SO+ZO), and random initialization combined with Zeroth-Order (ZO) optimization (Random+ZO), respectively.}
\centering
\resizebox{0.8\linewidth}{!}{
\begin{tabular}{ccc|cc|cc|cc|cc|cc}
\toprule[1.2pt]
\multirow{2}{*}{} & \multicolumn{6}{c|}{Dodging}                                                                             & \multicolumn{6}{c}{Impersonation}                                                                       \\ \cline{1-13} 
                  & \multicolumn{2}{c|}{Random+ZO} & \multicolumn{2}{c|}{SO+ZO} & \multicolumn{2}{c|}{SO}                        & \multicolumn{2}{c|}{Random+ZO} & \multicolumn{2}{c|}{SO+ZO} & \multicolumn{2}{c}{SO}                        \\ \cline{1-13} 
                  & 65.79\%      & 1972      & 97.73\%        & 834        &96.65\%        & 11                      & 50.52\%       & 2874      & 87.37\%        & 1311       &  72.83\%                       &  27                     \\ \bottomrule[1.2pt]
\end{tabular}}
\label{tab:three}
\end{table*}

\subsubsection{Attacks in the physical world}
We show the results of our adversarial patch in the physical environment. We first perform simulated successful attacks on different subjects in the digital environment, and then conduct experiments in the physical world. 

To make digital simulation results better adapt to the physical environment, we process the smoothness of the perturbation pattern. Specifically, during each iteration of the pattern generation, we first obtain a pattern with half the size of the original patch by scaling down or averaging pooling, and then enlarge the image back to the original size by bilinear interpolation. The reduced pattern retains the key information of the pattern. After zooming, the smoothness of the image is improved, avoiding the problem that the pattern generated per pixel is sensitive to the position point when it is transferred to the physical environment. Even if the pasting position is slightly a few pixels away, the attack effect can still be preserved. We also try to use Total Variation (TV) \cite{glass2019} loss to enhance the smoothness, but the actual effect is not as good as the scaling process. When printing, we use photo paper rather than ordinary paper as the patch material to recreate the colors of the digital simulation as realistic as possible.

 {Before applying to the physical environment, we first verify the robustness to the translation and rotation changes in the digital world. We translate the patch by 2, 3, and 5 pixels, and rotate it by 5 and 15 degrees to test the performance of the attack. Table \ref{tab:trans} lists the attack success rate (ASR) of dodging attacks on FaceNet and ArcFace50 models.
It can be seen that for translation, moving 2 pixels has little effect on the attack (ASR for FaceNe does not drop, ArcFace50 drops 0.72\%). Even with a 5-pixel shift, the ASR of 79.24\% and 53.11\% can still be achieved on two models.
For rotation, within 5 degrees of rotation, the ASR of the two models decreases slightly by 2.96\% and 8.23\%, respectively, which can meet our requirements for applying to the physical environment.
}

 {For the performance in the physical environment}, we record the video when faces are moving within a small range of the current posture, and count the frame proportion of successful physical attacks  as the attack success rate.
Table \ref{tab:front} shows the results of the frontal face.
To test the performance under various physical conditions, we further change the face posture, the distance from the camera and the illumination. For face postures, we take the conditions of the frontal face, yaw angle rotation of $\pm 25^{\circ}$ (the mean value at $25^{\circ}$ and $- 25^{\circ}$), $\pm 45^{\circ}$ and pitch angle rotation of $\pm 15^{\circ}$. The results are shown in Table \ref{tab:env}. 

It can be seen from Table \ref{tab:front} that our method maintains high physical ASR (100.0\% and 83.33\%) on FaceNet, and the results are also good on ArcFace34 and CosFace50. Although there is a slight decline in commercial API, the ratio of 33.16\% and 25.23\% of successful frames is enough to bring potential risks to commercial applications.
In Table \ref{tab:env}, when the pose changes in a small range ($\pm \ y 25^{\circ}$, $\pm \ p15^{\circ}$), ASR still maintains a high value. Even if the deflection angle is slightly larger ($\pm \ y45^{\circ}$), it can still maintain an average of 31.13\%. 
The effect of distance is small, and when lighting is changed, the results are still at an acceptable level. When performing impersonation attacks under different conditions, the face can be recognized as a false identity different from the true identity, but not the target identity, which may lead to a slightly lower result.
Visual results are shown in Figure \ref{fig:phy}.



\subsection{Combined with gradient estimation attacks}\label{sec:extend1}


 {The results of combining our simultaneous optimization with the gradient estimation attack are given in Table \ref{tab:three}, where SO denotes the results output by our simultaneous optimization method. SO+ZO denotes the combined version of our SO method and Zeroth-Order (ZO) attack, i.e., using the adversarial patch of SO as the initialized values of ZO. Random+ZO denotes the ZO attack with random initialized values of ZO. From the table, we can see that SO+ZO outperforms random+ZO significantly. For the dodging attack, the ASR increases from 65.79\% to 97.73\%, and NQ decreases from 1972 to 834. For the impersonation attack, the ASR increases from 50.52\% to 87.37\%, and NQ decreases from 2874 to 1311. This contrast proves that our SO method indeed provides a good initialization for the gradient estimation attack, reducing their high query costs and improving their attack
success.  In addition, we also see SO+ZO obtains better ASR than SO for both the dodging and impersonation, but introduces much more query consumption. }

\begin{figure*}[h]
\centering
\includegraphics[width=0.9\textwidth]{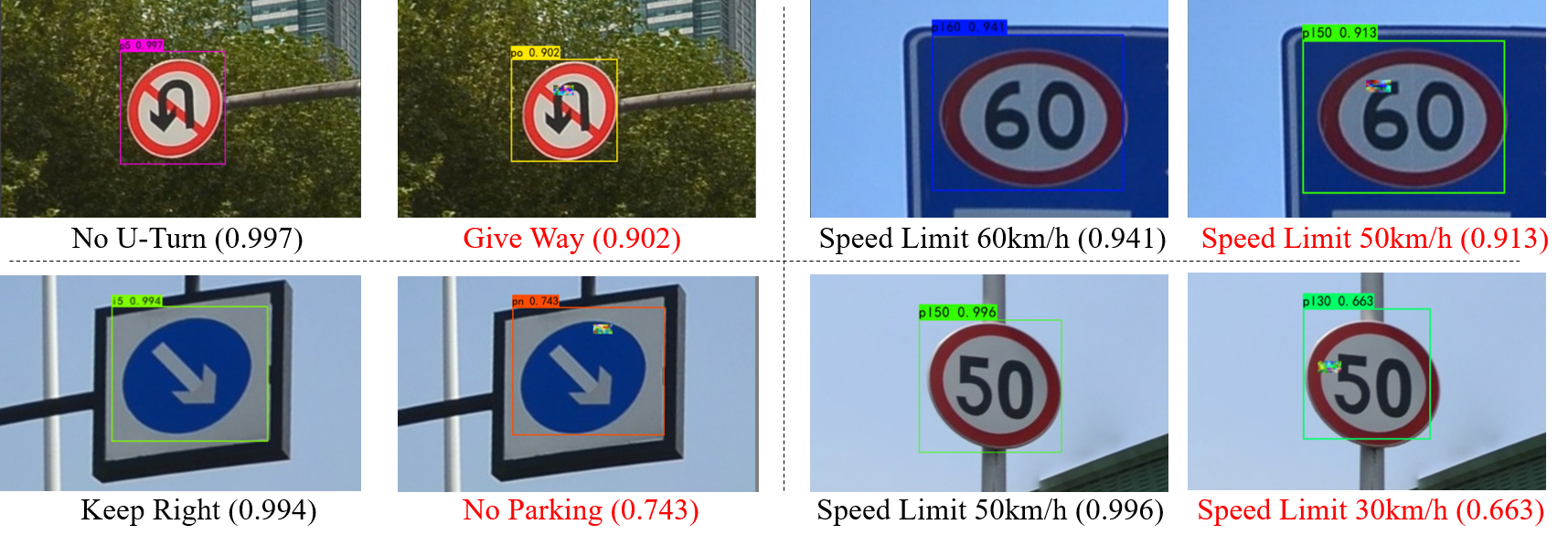}
\caption{The attack results on the traffic sign recognition model. The bounding boxes are marked on the images, and the text below the picture indicates the corresponding sign category and confidence score.}
\label{fig:sign}
\end{figure*}

\section{Extensions to traffic sign recognition }\label{sec:traffic}
We also use our method to attack the traffic sign recognition model to prove its wide applicability. We conduct experiments on the Tsinghua-Tencent 100K
(TT100K) dataset \cite{ttk100} which contains many different types of Chinese traffic signs, covering a variety of weather and light conditions.
Specifically, we use YOLOv4 \cite{yolov4} and YOLOv3 \cite{yolov3} as surrogate models, and NanoDet\footnote{https://github.com/topics/nanodet} \cite{=nanodet} as the target model to be attacked. We choose NanoDet because it can run on the mobile phone, which can simulate the vehicle perspective.
The size of the adversarial patch is set as $30 \times 15$, and the allowable patch sticking region is restricted within the detection bounding box of the traffic sign predicted in the clean image. The rest of the attack settings are consistent with the face recognition task.

 {The attack success rate (ASR) and the number of queries (NQ) of untargeted and targeted attacks are shown in Table \ref{tab:nano}. It can be seen that our attack can achieve ASR of 68.57\% and 45.62\% in untargeted and targeted attacks with only a few queries, respectively, which is similar to the face recognition task. This performance demonstrates the effectiveness of our method in the traffic sign recognition task, showing its generalization ability. }

\begin{table}[t]
\caption{ {The attack success rate (ASR) and the number of queries (NQ) for untargeted and targeted attacks on NanoDet.}}
\centering 
\resizebox{0.7\linewidth}{!}{
\begin{tabular}{c|cc|cc}
\toprule[1.2pt]
\multirow{2}{*}{Target model} & \multicolumn{2}{c|}{Untargeted}   & \multicolumn{2}{c}{Targeted}      \\ \cline{2-5} 
                              & \multicolumn{1}{l|}{ASR}     & NQ & \multicolumn{1}{l|}{ASR}     & NQ \\ \hline
NanoDet                       & \multicolumn{1}{l|}{68.57\%} & 36 & \multicolumn{1}{l|}{45.62\%} & 45 \\ \bottomrule[1.2pt]
\end{tabular}\label{tab:nano}}
\end{table}

Several  {visual} examples are shown in Figure \ref{fig:sign}. The bounding boxes are marked on the image, and the predicted category and confidence score are given below the image.
In these four groups, the image on the left represents a clean example that has not been attacked, which can be correctly detected and classified by the target detection model. The picture on the right shows the corresponding adversarial example.
For example, No U-Turn is recognized as Giving Way, and Keeping Right is recognized as No Parking.
After being attacked by the adversarial patch, the traffic sign is still detected, but the prediction result of its category is wrong, even though its semantics does not change for humans.

\section{Conclusion}\label{sec:conclusion}
In this paper, we proposed an efficient method to achieve the simultaneous optimization of the positions and perturbations to create an adversarial patch  {in the black-box setting.}
This process was formulated into a reinforcement learning framework.  {Through the design of the agent and different policy functions, the attack variables involving the patch's position and perturbation can be simultaneously realized  with a few query times.}
Extensive experiments demonstrated that our method could effectively improve the attack effect, and experiments on the commercial FR API and physical environments confirmed that it had the practical value and could be used to help evaluate the robustness of DNN models in real applications.  Besides, our method can also be adapted to other applications, such as  {autonomous} driving, etc.

\section*{Acknowledgment}
This work is supported by National Key R$\&$D Program of China (Grant No.2020AAA0104002), National Natural Science Foundation of China (No.62076018). We also thank anonymous reviewers for their valuable suggestions. 

\ifCLASSOPTIONcaptionsoff
  \newpage
\fi

\bibliographystyle{IEEEtran}
\bibliography{arxiv}

\begin{IEEEbiography}[{\includegraphics[width=1in,height=1.25in,clip,keepaspectratio]{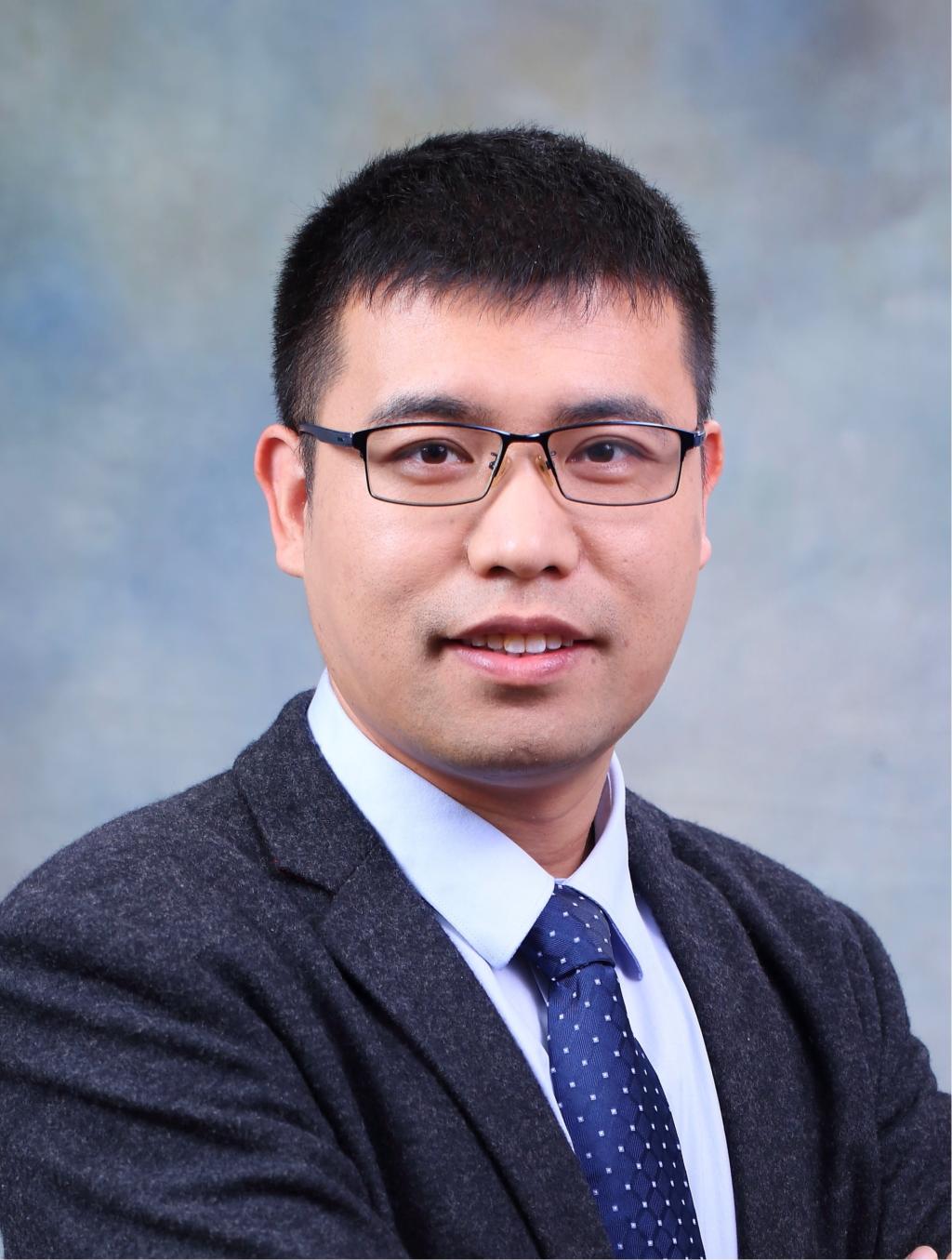}}]{Xingxing Wei}
received his Ph.D degree in computer science from Tianjin University, and B.S. degree in Automation from Beihang University, China. He is now an Associate Professor in Beihang University (BUAA). His research interests include computer vision, adversarial machine learning and its applications to multimedia content analysis. He is the author of referred journals and conferences in IEEE TPAMI, TMM, TCYB, TGRS, IJCV, PR, CVIU,  CVPR, ICCV, ECCV, ACMMM, AAAI, IJCAI etc.
\end{IEEEbiography}
\vspace{-1cm}
\begin{IEEEbiography}[{\includegraphics[width=1in,height=1.25in,clip,keepaspectratio]{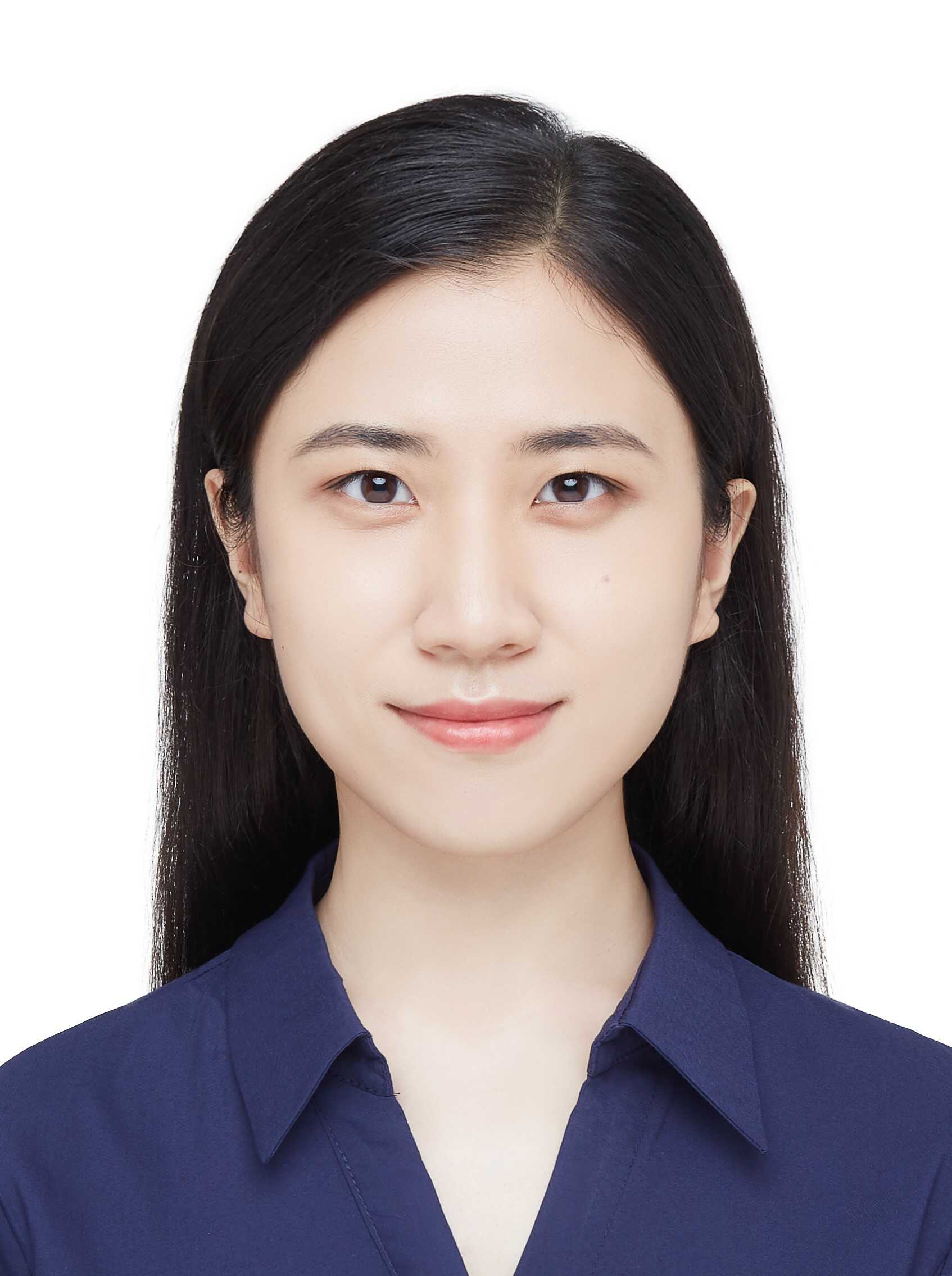}}]{Ying Guo} is now a Master student at School of Computer Science and Engineering, Beihang University (BUAA).
Her research interests include deep learning and adversarial robustness in machine learning.
\end{IEEEbiography}
\vspace{-1cm}
\begin{IEEEbiography}[{\includegraphics[width=1in,height=1.25in,clip,keepaspectratio]{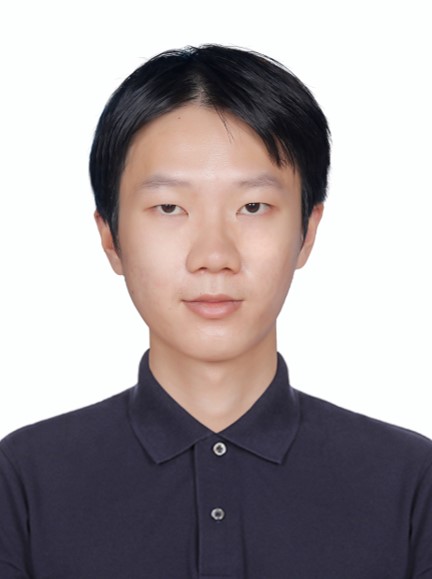}}]{Jie Yu} is now a Master student at School of Computer Science and Engineering, Beihang University (BUAA).
His research interests include deep learning, compter vision and adversarial robustness.
\end{IEEEbiography}
~\\
~\\
\textbf{Bo Zhang} is the senior researcher with the Tencent Corporation in Shenzhen, China. His research interest is AI security.

\end{document}